\documentclass[11pt]{article}

\usepackage{arxiv}
\usepackage[utf8]{inputenc}
\usepackage[T1]{fontenc}
\usepackage{hyperref}
\usepackage{url}
\usepackage{booktabs}
\usepackage{amsfonts}
\usepackage{amsmath}
\usepackage{amssymb}
\usepackage{nicefrac}
\usepackage{graphicx}
\usepackage{xcolor}
\usepackage{multirow}
\usepackage{subcaption}
\usepackage{enumitem}
\usepackage{float}
\usepackage{needspace}
\usepackage{cleveref}
\usepackage[authoryear,round]{natbib}
\usepackage{tikz}
\usetikzlibrary{positioning,arrows.meta,shapes.geometric,fit,calc,backgrounds}
\usepackage{pgfplots}
\pgfplotsset{compat=1.18}
\usepgfplotslibrary{groupplots}
\usepgfplotslibrary{fillbetween}

\hypersetup{colorlinks=true, linkcolor=blue!50!black, citecolor=blue!50!black, urlcolor=blue!50!black}
\definecolor{cGnd}{HTML}{2A78D6}    
\definecolor{cGndDk}{HTML}{1B5A9E}  
\definecolor{cGnd2}{HTML}{0E8F8F}   
\definecolor{cUng}{HTML}{EB6834}    
\definecolor{cUngDk}{HTML}{D1495B}  
\definecolor{cUng2}{HTML}{B01743}   
\definecolor{cStu}{HTML}{4A3AA7}    
\definecolor{cInk}{HTML}{262626}    
\definecolor{cMut}{HTML}{6E6E6E}    
\definecolor{cGrid}{HTML}{DDDDDD}   

\pgfplotsset{
  every axis/.append style={
    font=\small,
    axis line style={draw=cMut!70, line width=0.5pt},
    tick style={draw=cMut!70, line width=0.5pt},
    label style={font=\footnotesize, color=cInk},
    tick label style={font=\footnotesize, color=cMut},
    major grid style={solid, cGrid, line width=0.4pt},
    legend style={font=\footnotesize, draw=none, fill=white, fill opacity=0.85, text opacity=1},
  },
  paperbar/.style={
    axis x line*=bottom, axis y line*=left,
  },
  swatch/.style={legend image code/.code={%
    \draw[##1] (0cm,-0.09cm) rectangle (0.28cm,0.14cm);}},
  lineswatch/.style={legend image code/.code={%
    \draw[##1] (0cm,0.03cm) -- (0.42cm,0.03cm);}},
}

\title{Interaction Scaling:\texorpdfstring{\\}{ }Grounding the Third Axis of Test-Time Compute}

\author{%
  \begin{tabular}{@{}c@{\hspace{4em}}c@{}}
    Bojie Li & Noah Shi \\
    Pine AI & University of Washington
  \end{tabular}%
}
\date{}
\runningtitle{Interaction Scaling: Grounding the Third Axis of Test-Time Compute}

\begin{document}
\maketitle

\begin{abstract}
There are two standard ways to spend more compute at test time: let a model \emph{reason} longer, or \emph{sample} more attempts and keep one. Both share a hidden limit: they are \emph{internal}. Every extra token comes from the same frozen weights and the same prompt, so neither can tell the model anything it does not already know. We study a third way, \emph{interaction}: the model proposes an artifact, an external instrument observes how it actually behaves, and the model revises. Each cycle imports a real observation, so interaction breaks through the ceiling the other two hit.

We argue that a single variable governs this third axis, \textbf{grounding}, and that it must hold on \emph{both} sides of the loop. The feedback that drives revision must come from an instrument that actually observes the flaw, and so must the metric that scores the result. On hard coding tasks at a fixed token budget, reasoning-only and best-of-$N$ sampling both plateau (the latter even when an oracle picks the best sample), while every interaction strategy keeps improving; our proposer--reviewer harness reaches a perfect $100\%$ pass rate with no run-to-run variance, and the gain holds across three model families. On rendered visual artifacts, the usual judge (a vision--language model, or VLM, reading a screenshot) rates $14$ of $15$ visibly broken figures ``perfect,'' because the screenshot hides the flaws before the judge can see them. A tool that measures the real layout instead shows the loop removing $40$--$74\%$ of defects across four modalities; and that same VLM, used as the \emph{reviewer}, makes slide layouts worse where the measuring tool repairs them. Interaction scaling is real and distinct from reasoning and sampling, but only visible when both the feedback and the metric are grounded.
\end{abstract}
\begin{center}
\footnotesize
Code: \url{https://github.com/19PINE-AI/interaction-scaling}\\[1pt]
Website: \url{https://01.me/research/interaction-scaling}
\end{center}
\vspace{-0.4em}

\begin{figure}[H]
\centering
\resizebox{0.72\linewidth}{!}{%
\begin{tikzpicture}[font=\small,
  box/.style={rounded corners=2pt,draw=cMut,thick,minimum height=8mm,align=center,fill=white,inner sep=4pt},
  prop/.style={box,fill=cStu!8,draw=cStu!70},
  env/.style={box,fill=black!4},
  gnd/.style={box,fill=cGnd!10,draw=cGnd},
  ung/.style={box,fill=cUng!8,draw=cUng},
  ar/.style={-{Latex[length=2mm]},thick,cMut},
  gar/.style={-{Latex[length=2mm]},very thick,cGnd},
  uar/.style={-{Latex[length=2mm]},thick,cUng,dashed}]
  \node[prop] (prop) {Proposer\\\footnotesize(frozen LLM)};
  \node[env, right=14mm of prop] (art) {Artifact\\\footnotesize code / HTML / SVG};
  \node[gnd, right=15mm of art, yshift=9mm] (inst)
     {Instrument\\\footnotesize\emph{execute / render / measure}};
  \node[gnd, right=13mm of inst] (score) {\footnotesize\textbf{(2)} grounded\\\footnotesize evaluation};
  \node[prop, below=7mm of inst, xshift=6mm] (rev) {Reviewer};
  \node[ung, below=15mm of art, xshift=3mm] (vlm) {VLM reads a \emph{screenshot}};
  \node[ung, right=10mm of vlm] (uscore) {\footnotesize blind feedback\\\footnotesize \& blind score};
  \draw[ar] (prop) -- node[above,font=\footnotesize]{generate} (art);
  \draw[ar] (art.east) -- (inst.west);
  \draw[gar] (inst) -- (score);
  \draw[gar] (inst.south) -- node[left,font=\footnotesize,cGnd,align=center,pos=0.45]{\textbf{(1)}\\defects} (rev.north);
  \draw[ar] (rev.south) |- ([yshift=-9mm]art.south) -| node[pos=0.25,below,font=\footnotesize]{targeted revision} (prop);
  \draw[uar] (art.south) -- (vlm.north);
  \draw[uar] (vlm) -- (uscore);
  \begin{scope}[on background layer]
    \node[draw=cGnd!60,dashed,rounded corners,fill=cGnd!4,fit=(inst)(score)(rev),inner sep=3mm,
      label={[cGnd,font=\footnotesize\bfseries]above:GROUNDED: measure the rendered DOM / run \texttt{pytest}}] {};
    \node[draw=cUng!60,dashed,rounded corners,fill=cUng!4,fit=(vlm)(uscore),inner sep=2.5mm,
      label={[cUng!85!black,font=\footnotesize\bfseries]below:UNGROUNDED: overflow cropped off-frame, rates broken figures ``perfect''}] {};
  \end{scope}
\end{tikzpicture}}
\caption{\textbf{Grounding must hold on both sides of the interaction loop.} One instrument observation is both \textbf{(1)} grounded \emph{feedback} (the defect list that drives revision and escapes the internal ceiling) and \textbf{(2)} grounded \emph{evaluation}, the score that makes the gain measurable. The default VLM-on-a-screenshot judge (orange lane) breaks both: the screenshot drops the defects before the model sees them.}
\label{fig:concept}
\end{figure}
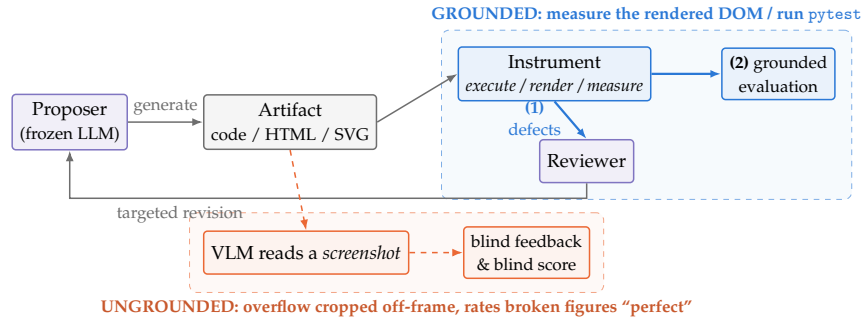

\clearpage
\section{Introduction}
\label{sec:intro}

Once a model is trained, the way we make it better on a hard task is no longer to add parameters or data (\emph{training-time scaling}) but to spend more compute per query on the frozen model: \emph{inference-time scaling} \citep{openai2024o1,snell2024scaling}. On the artifact-producing tasks we study (code, web pages, slides, figures, animations, video edits, research reports), this is the lever we can still pull. Prior work formalizes two ways to pull it. \emph{Reasoning scaling} \citep{wei2022chain,yao2023tree,deepseek2025r1,snell2024scaling} spends more tokens thinking before committing; \emph{sampling scaling} \citep{wang2023selfconsistency,brown2024large} draws more attempts and selects one. They look different, but share a property this paper treats as fundamental: both are \textbf{internal}. Every extra token, whether in a longer chain of thought or another sample, comes from the same frozen weights and the same fixed prompt. More internal compute reshuffles what the model already has; it imports nothing new.

We study a third form of inference-time scaling that breaks out of this closed loop: \emph{interaction scaling}, in which the model queries an external instrument that observes the artifact itself. The model runs the tests, renders the page and measures the layout, or issues the search query and reads the result. Such an observation is not a function of the weights; it reports how the artifact actually behaves, and can carry information the model never had. This axis is well established in practice \citep{shinn2023reflexion,yao2023react,gou2024critic,shen2025thinking}, but accounts of \emph{why} and \emph{when} it beats internal scaling remain informal and, we argue, only half-stated. Industry practice shows the stakes: in a controlled $900$-run deployment study at ByteDance, frontier coding models passed the functional-correctness bar yet fell short on delivery quality until a feedback ``harness'' was added, and the study had to invent a multi-axis quality metric before the gap was even visible \citep{hong2026aicoding}. That is a field sighting of exactly the two-sided problem we formalize below (\Cref{sec:related}).

\paragraph{Grounding is the variable, on both sides of the loop.} Our thesis is that one variable governs interaction scaling: \textbf{grounding}. Feedback is \emph{grounded} when it comes from an instrument observing the artifact's actual form or behavior, not from a model voicing an opinion about it. And grounding must hold on \emph{both} sides of the proposer--reviewer loop (\Cref{fig:concept}):
\begin{enumerate}[leftmargin=1.4em,itemsep=1pt,topsep=2pt]
\item \textbf{Grounded feedback} (the signal that drives revision): run the code and read the failing test, render the HTML and measure where each element lands, issue the query and read the result. Prior work focuses on this side; a short information argument (\Cref{sec:framework}) explains why it lets interaction break the internal ceiling.
\item \textbf{Grounded evaluation} (the metric that \emph{measures} the gain): the same observation, used as the score. This side is almost always overlooked, and getting it wrong hides the effect entirely.
\end{enumerate}
The second point matters more than it may seem. The standard evaluator for visual artifacts is a vision--language model (VLM) reading a screenshot, and a screenshot \emph{drops the flaw before the judge ever sees it}: overflowing content is cropped off-frame, and small overlaps fall below the model's visual acuity. On dense academic figures this judge rates $14$ of $15$ single-shot renders ``perfect,'' where a direct measurement finds only $3$ actually clean (\Cref{sec:grounded_eval}); used as the \emph{reviewer}, the same VLM makes slide layouts worse (\Cref{sec:feedback}). Ungrounded on either side, the third axis either does not fire or cannot be seen.

\begin{figure}[t]
\centering
\resizebox{\linewidth}{!}{%
\begin{tikzpicture}
\begin{axis}[paperbar,name=axhcode,width=4.9cm,height=4.8cm,
  ybar,bar width=9pt,ymin=0,ymax=118,
  xlabel={\footnotesize (a) code: pass rate (\%)},
  symbolic x coords={dev,deepspec},xtick=data,
  xticklabels={hard suite,deep-spec},
  x tick label style={font=\footnotesize,color=cMut},
  enlarge x limits=0.45,ytick={0,25,50,75,100},ymajorgrids,
  legend style={at={(0.5,1.02)},anchor=south,legend columns=2,
    /tikz/every odd column/.append style={column sep=2pt}}]
  \addplot[swatch,draw=none,fill=black!22]
    coordinates {(dev,66.7) (deepspec,69.7)};
    \addlegendentry{single-shot}
  \addplot[swatch,draw=none,fill=cGnd!80]
    coordinates {(dev,100) (deepspec,100)};
    \addlegendentry{reviewed}
  \node[font=\scriptsize\bfseries,cInk,anchor=south] at (axis cs:dev,101) {\hspace{11pt}100};
  \node[font=\scriptsize\bfseries,cInk,anchor=south] at (axis cs:deepspec,101) {\hspace{11pt}100};
  \node[font=\scriptsize,cMut,anchor=south] at (axis cs:dev,67.5) {\hspace{-15pt}66.7};
  \node[font=\scriptsize,cMut,anchor=south] at (axis cs:deepspec,70.5) {\hspace{-15pt}69.7};
\end{axis}
\begin{axis}[paperbar,at={($(axhcode.east)+(1.5cm,0)$)},anchor=west,
  width=7.0cm,height=4.8cm,
  ybar,bar width=11pt,ymin=0,ymax=100,
  xlabel={\footnotesize (b) visual: layout defects removed (\%)},
  symbolic x coords={Figures,Slides,Web,Anim.},xtick=data,
  x tick label style={font=\footnotesize,color=cMut},
  enlarge x limits=0.18,ytick={0,25,50,75,100},ymajorgrids]
  \addplot[draw=none,fill=cGnd!80,bar shift=0pt]
    coordinates {(Figures,74) (Slides,73) (Web,47) (Anim.,40)};
  \node[font=\scriptsize\bfseries,cInk,anchor=south] at (axis cs:Figures,75) {$-$74\%};
  \node[font=\scriptsize\bfseries,cInk,anchor=south] at (axis cs:Slides,74) {$-$73\%};
  \node[font=\scriptsize\bfseries,cInk,anchor=south] at (axis cs:Web,48) {$-$47\%};
  \node[font=\scriptsize\bfseries,cInk,anchor=south] at (axis cs:Anim.,41) {$-$40\%};
\end{axis}
\begin{axis}[paperbar,at={($(axhcode.east)+(9.9cm,0)$)},anchor=west,
  width=4.3cm,height=4.8cm,
  ybar,bar width=9pt,ymin=0,ymax=1.18,clip=false,
  xlabel={\footnotesize (c) saturated: score in $[0,1]$},
  symbolic x coords={Video,Research},xtick=data,
  x tick label style={font=\footnotesize,color=cMut},
  enlarge x limits=0.45,ytick={0,0.25,0.5,0.75,1.0},ymajorgrids]
  \draw[cMut,densely dotted] (rel axis cs:0,0.847)--(rel axis cs:1,0.847);
  \node[font=\scriptsize\itshape,cMut,anchor=south west,inner sep=1pt] at (rel axis cs:0.02,0.85) {ceiling};
  \addplot[draw=none,fill=black!22]
    coordinates {(Video,0.70) (Research,0.98)};
  \addplot[draw=none,fill=cGnd!80]
    coordinates {(Video,0.74)};
  \node[font=\scriptsize,cMut,anchor=south] at (axis cs:Video,0.75) {\hspace{11pt}0.74 (n.s.)};
  \node[font=\scriptsize,cMut,anchor=south,fill=white,inner sep=1pt] at (axis cs:Research,0.99) {\hspace{-9pt}0.98};
\end{axis}
\end{tikzpicture}}
\caption{\textbf{Results at a glance: one grounded loop, seven modalities.} \textbf{(a)}~Execution feedback lifts both code suites to a strict $100\%$ pass rate, recovering every single-shot failure (3-seed means; \Cref{sec:interaction}). \textbf{(b)}~Grounded geometry feedback removes $40$--$74\%$ of the layout defects a deterministic DOM instrument measures, on all four visual modalities (every reduction statistically decisive; \Cref{sec:grounded_eval}). \textbf{(c)}~The two remaining modalities are scoped negatives: video editing is already strong single-shot (the reviewed lift is not significant), and deep research saturates at $0.98$ single-shot factual accuracy, leaving no headroom for feedback to claim (\Cref{sec:grounded_eval}).}
\label{fig:headline}
\end{figure}
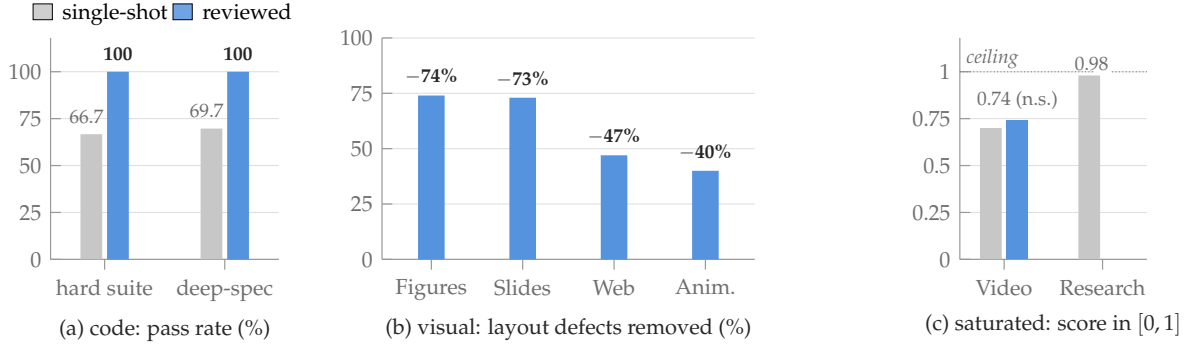

\begin{figure}[t]
\centering
\begin{subfigure}[b]{0.49\linewidth}
\centering
\begin{tikzpicture}
\node[anchor=south west,inner sep=0] (img) {\includegraphics[width=\linewidth]{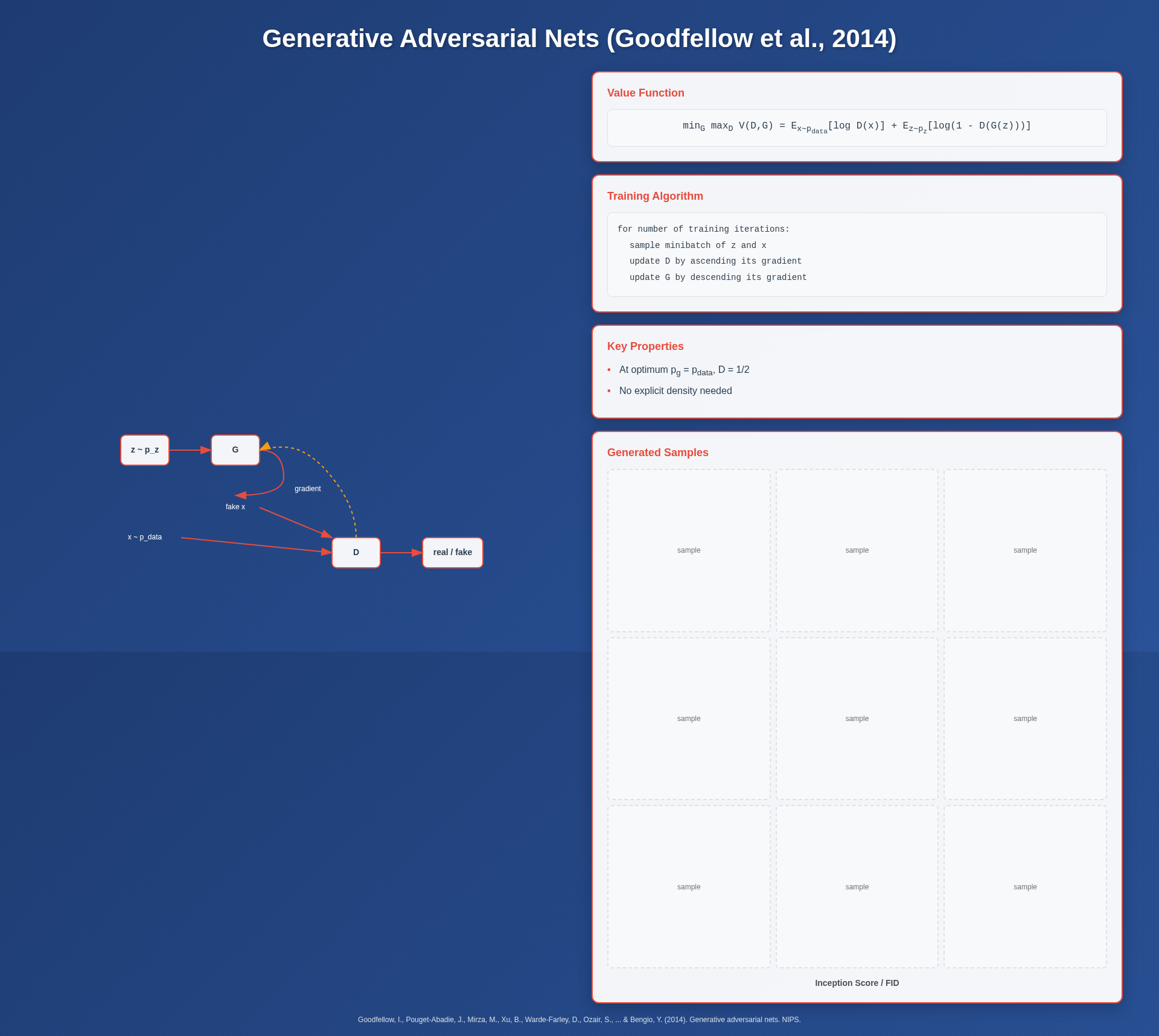}};
\draw[cUngDk,dashed,line width=0.9pt] ($(img.south west)!0.371!(img.north west)$) -- ($(img.south east)!0.371!(img.north east)$);
\node[anchor=north east,font=\tiny\bfseries,text=cUngDk,fill=white,fill opacity=0.85,text opacity=1,inner sep=2pt]
  at ($(img.south east)!0.371!(img.north east)$) {$16{:}9$ slide frame ends here};
\end{tikzpicture}
\caption{\textbf{Single-shot} ($6$ measured defects): the content overflows the $16{:}9$ slide frame (dashed line) by more than half the frame's height: exactly the defect class a screenshot judge never sees, because the capture crops at the frame.}
\label{fig:ba-ss}
\end{subfigure}\hfill
\begin{subfigure}[b]{0.49\linewidth}
\centering
\includegraphics[width=\linewidth]{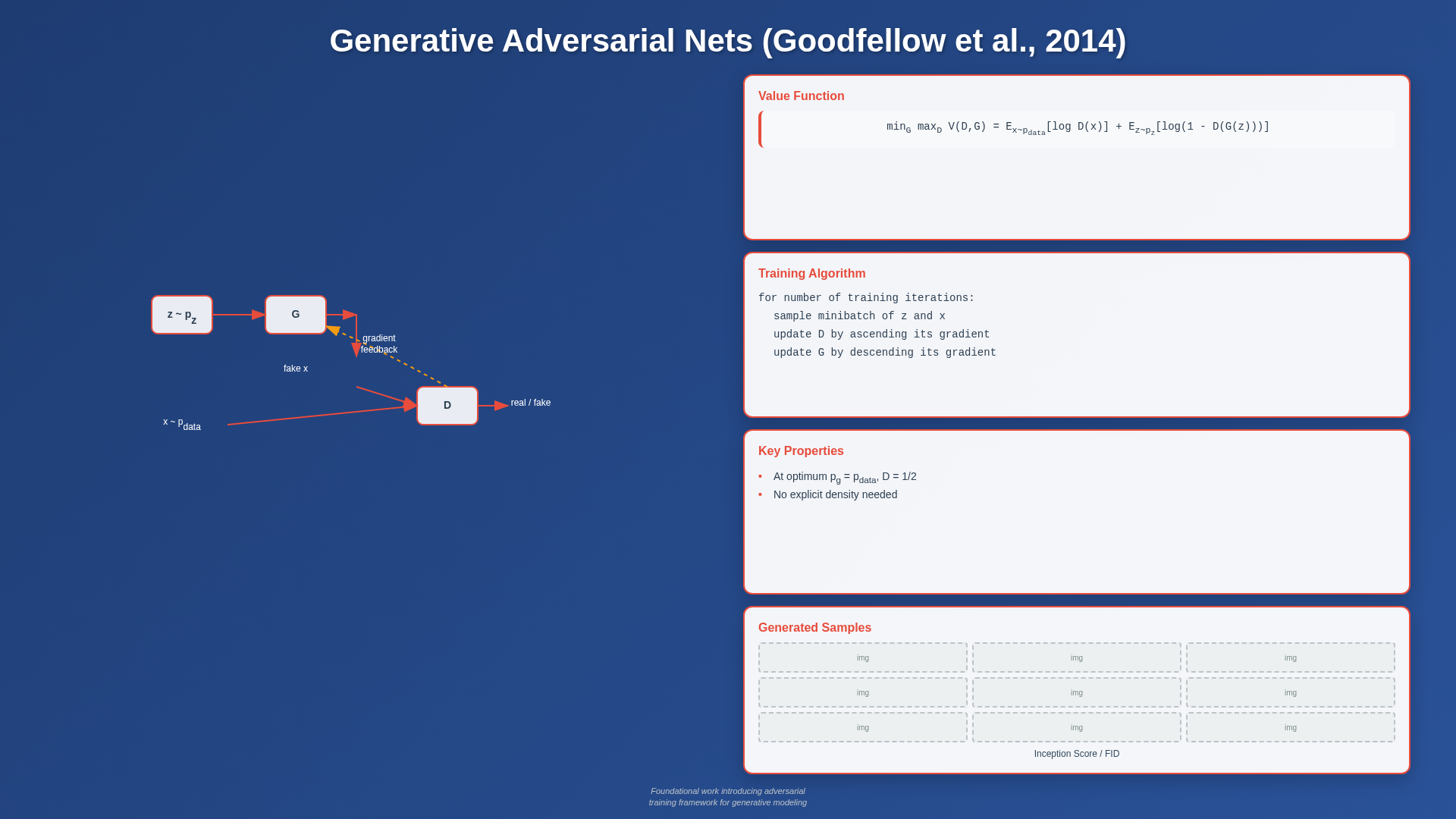}
\caption{\textbf{After grounded feedback} ($0$ defects): the same task after the propose$\to$measure$\to$revise loop: the sample grid is compacted and the panels rebalanced so every element sits inside the frame.}
\label{fig:ba-rv}
\end{subfigure}
\caption{\textbf{A concrete example of what the loop fixes.} A dense real-paper slide task (the GAN paper), rendered full-page at equal width: single-shot (\subref{fig:ba-ss}) vs.\ reviewed (\subref{fig:ba-rv}). The deterministic geometry instrument counts $6$ defects single-shot and $0$ after review (the same $6\to0$ reduction replicates in a second seed), and a screenshot judge sees none of them (\Cref{sec:grounded_eval}).}
\label{fig:beforeafter}
\end{figure}

\paragraph{What the loop delivers, at a glance.} \Cref{fig:headline} summarizes what one frozen model in one harness achieves across the seven modalities we test. Grounded execution feedback lifts both hard code suites to a perfect $100\%$ pass rate; grounded geometry feedback removes $40$--$74\%$ of the measured layout defects on four visual modalities; and the two modalities that leave no single-shot \emph{headroom} (no room to improve before the metric's ceiling) are reported as honest negatives, not manufactured wins. \Cref{fig:beforeafter} makes this concrete on one dense-slide task: the defects the loop removes are exactly the ones the standard screenshot judge cannot see.

\Needspace*{6\baselineskip}
\paragraph{Contributions.}
\begin{enumerate}[leftmargin=1.4em,itemsep=1pt,topsep=2pt]
\item \textbf{A grounding framework (\Cref{sec:framework}).} We sort feedback into two kinds (\emph{ungrounded} model opinion vs.\ \emph{grounded} instrument observation) and add a \emph{coverage} principle: grounded feedback helps exactly as far as its instrument can observe. A short information argument (formalized in \Cref{app:ceiling}) explains why internal scaling saturates, and the same argument shows that an ungrounded \emph{metric} cannot measure the gain. Beyond the known \emph{biases} of model-as-judge evaluation \citep{zheng2023judging}, we pin down a sharper failure: where quality is a measurable physical property, a screenshot judge is \emph{structurally blind}, not merely noisy.
\item \textbf{Interaction keeps scaling where internal compute stops (\Cref{sec:interaction}).} At the same token budget, reasoning-only and best-of-$N$ saturate (the latter even with an oracle verifier), while every interaction strategy climbs toward a perfect pass rate; the proposer--reviewer harness does so at the lowest token cost and with zero seed variance.
\item \textbf{The feedback instrument must observe the defect (\Cref{sec:feedback}).} Swapping only the reviewer's instrument on a fixed suite: on behavioral bugs, execution feedback converges far more cheaply than critique; a linter (grounded, but observing only surface form) buys nothing; and a screenshot-reading VLM reviewer makes slide geometry \emph{worse}, where a geometry reviewer repairs it.
\item \textbf{So must the metric (\Cref{sec:grounded_eval}).} The standard VLM judge passes $14$ of $15$ broken figures; a tool that measures the rendered layout instead reveals large, statistically decisive defect reductions across four visual modalities, gains the standard metric cannot see.
\end{enumerate}

\section{A Framework for Internal and External Test-Time Compute}
\label{sec:framework}

\subsection{Internal vs.\ external test-time compute}
\label{sec:internal-external}

All three axes we consider are forms of \emph{inference-time scaling}: on a frozen model, they spend more compute per query rather than adding parameters or data \citep{openai2024o1,snell2024scaling}. What separates them is where the extra compute draws its information from. Reasoning and sampling differ in mechanics, but not in kind. A longer chain of thought re-derives consequences of what the weights and prompt already contain; best-of-$N$ re-draws from the same distribution and picks one. In both, every token comes from the same two sources (frozen weights, fixed prompt), so we call them \textbf{internal} scaling. \textbf{External} scaling is different in kind: the artifact is handed to an instrument outside the model (a test runner, a layout engine, a search index), and the instrument's observation is fed back into the next generation. That observation reflects the \emph{artifact's actual behavior}, not the model's beliefs about it.

\subsection{A feedback taxonomy, and the coverage principle}
\label{sec:taxonomy}

\begin{figure}[t]
\centering
\resizebox{\linewidth}{!}{%
\begin{tikzpicture}[font=\small,
  ubox/.style={rounded corners=2pt,draw=cUng,fill=cUng!8,minimum height=10mm,align=center,inner sep=5pt},
  gbox/.style={rounded corners=2pt,draw=cGnd,fill=cGnd!8,minimum height=10mm,align=center,inner sep=5pt},
  lab/.style={font=\footnotesize\bfseries}]
  \node[ubox] (crit) at (0,0) {LLM critique\\\footnotesize re-reads its own output};
  \node[ubox, right=5mm of crit] (vlmshot) {VLM on a screenshot\\\footnotesize channel drops the defect};
  \node[font=\footnotesize\itshape,cUng!85!black,right=8mm of vlmshot,align=left]
    {bounded by what the model already knows\\(the internal ceiling, \Cref{sec:ceiling})};
  \node[gbox] (lint) at (0,-3.1) {static form\\\footnotesize linter, type checker};
  \node[gbox, right=4mm of lint] (exec) {execution\\\footnotesize tests, tracebacks};
  \node[gbox, right=4mm of exec] (geom) {rendered geometry\\\footnotesize DOM bounding boxes};
  \node[gbox, right=4mm of geom] (temp) {temporal\\\footnotesize per-frame geometry};
  \node[gbox, right=4mm of temp] (fact) {external facts\\\footnotesize search verification};
  \draw[-{Latex[length=2.2mm]},thick,cGnd!70!black]
    ([yshift=-3.5mm]lint.south west) -- ([yshift=-3.5mm]fact.south east)
    node[midway,below=2pt,font=\footnotesize,cGnd!70!black]
    {coverage, what the instrument can observe: \ form $\rightarrow$ behavior $\rightarrow$ world};
  \begin{scope}[on background layer]
    \node[draw=cUng!60,dashed,rounded corners,fill=cUng!3,fit=(crit)(vlmshot),inner sep=2.5mm,
      label={[cUng!85!black,lab]above:UNGROUNDED: a model's opinion}] (ug) {};
    \node[draw=cGnd!60,dashed,rounded corners,fill=cGnd!3,fit=(lint)(fact),inner sep=2.5mm,yshift=-2mm,
      label={[cGnd,lab]above:{GROUNDED: an instrument's observation \normalfont\itshape imports new information, as far as it can observe}}] (gd) {};
  \end{scope}
\end{tikzpicture}}
\caption{\textbf{The feedback taxonomy.} Feedback is either a model's opinion (\emph{ungrounded}, top) or an instrument's observation of the artifact (\emph{grounded}, bottom); grounded subtypes are named by what the instrument observes, and the \emph{coverage} axis orders them by reach. The screenshot-fed VLM sits in the ungrounded lane because the signal that enters the loop is a model's reading of an already-lossy view. Ungrounded feedback is subject to the internal ceiling (\Cref{sec:ceiling}); grounded feedback escapes it, but only for defects inside the instrument's coverage: a linter cannot see a runtime bug, and a screenshot cannot see off-canvas overflow.}
\label{fig:taxonomy}
\end{figure}
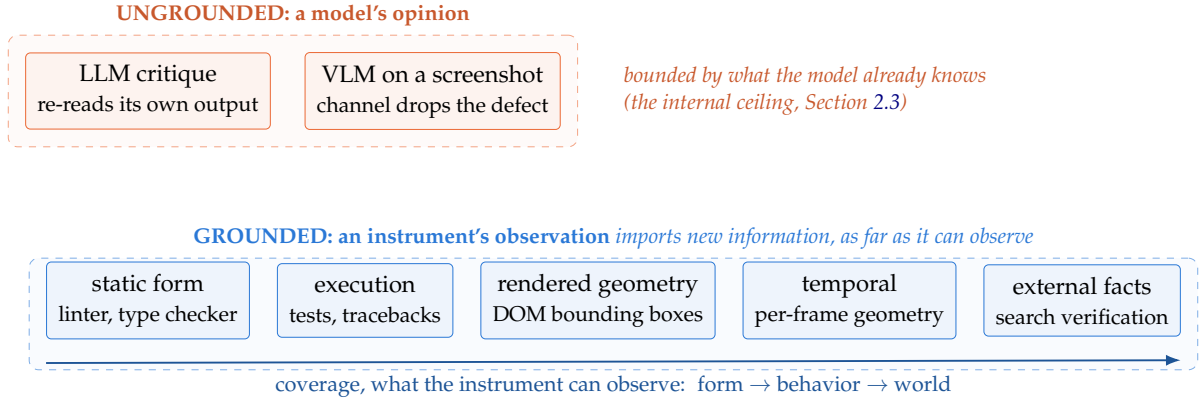

We classify a reviewer's feedback channel by \emph{who produces the signal} (\Cref{fig:taxonomy}):
\begin{itemize}[leftmargin=1.4em,itemsep=1pt,topsep=2pt]
\item \textbf{Ungrounded feedback} is a model's opinion: an LLM critiques its own artifact, or a VLM rates a screenshot. The screenshot case is worth care. A screenshot \emph{is} an instrument observation, but a lossy one that drops exactly the defects at issue (content off the canvas, overlaps too small to see), so the signal that enters the loop is the model's reading of those pixels. We classify a channel by what really produces its signal: here, the model.
\item \textbf{Grounded feedback} is an instrument's observation, its subtypes named by \emph{what the instrument observes}: \textbf{static form} (linters, type checkers inspect the source), \textbf{execution behavior} (tests report the failing assertion), \textbf{rendered geometry} (a layout engine reports every element's true bounding box), \textbf{temporal behavior} (the same, per animation frame), and \textbf{external facts} (a search engine checks a claim against the world).
\end{itemize}
Grounding alone, however, is not sufficient, and this is the second half of the taxonomy:

\begin{center}
\begin{tikzpicture}
\node[draw=cGnd!70,fill=cGnd!5,rounded corners=3pt,inner sep=8pt,text width=0.9\linewidth]{%
\textbf{The coverage principle.} A feedback channel helps exactly as far as its instrument's observational reach. A linter is grounded but observes only form, so it cannot see a runtime bug; a screenshot is an observation whose channel crops the defect out, so a VLM judging it cannot see broken geometry. The instrument must \emph{observe the property that is broken}.};
\end{tikzpicture}
\end{center}

The coverage principle is what unifies the paper's results: it predicts that execution feedback beats critique on behavioral bugs while a linter does not (\Cref{sec:feedback}), that a screenshot-fed VLM fails both as a reviewer (\Cref{sec:feedback}) and as a judge (\Cref{sec:grounded_eval}), and that a deterministic geometry instrument succeeds at both.

\subsection{Why internal scaling saturates}
\label{sec:ceiling}

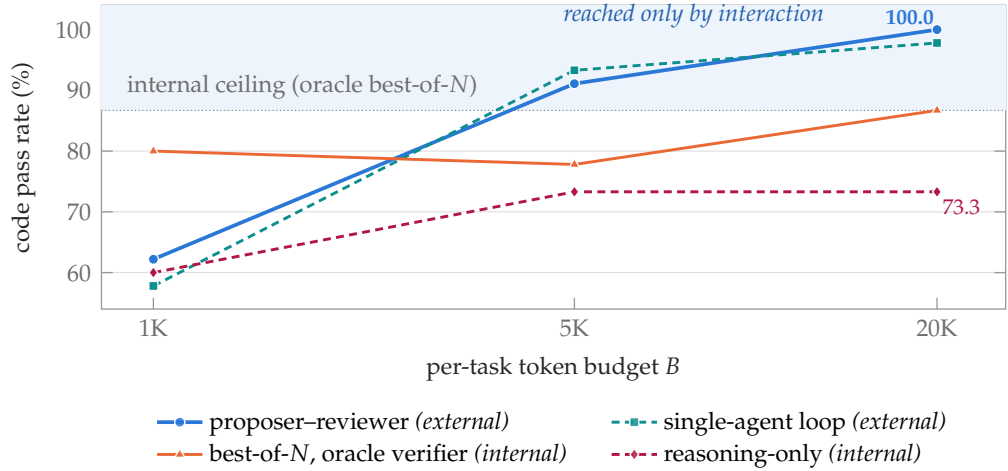
\begin{figure}[t]
\centering
\begin{tikzpicture}
\begin{axis}[width=0.86\linewidth,height=5.6cm,
  xmode=log,log basis x=10,
  xlabel={per-task token budget $B$},ylabel={code pass rate (\%)},
  xmin=820,xmax=26000,ymin=54,ymax=104,clip=false,
  xtick={1000,5000,20000},xticklabels={1K,5K,20K},
  ytick={60,70,80,90,100},ymajorgrids,
  legend style={at={(0.5,-0.30)},anchor=north,legend columns=2,
    /tikz/every even column/.append style={column sep=0.5cm}},legend cell align=left]
  \fill[cGnd!8] (axis cs:820,86.7) rectangle (axis cs:26000,104);
  \node[font=\footnotesize\itshape,cGnd!80!black,anchor=east] at (axis cs:13500,102.3)
     {reached only by interaction};
  \draw[cMut,densely dotted] (axis cs:820,86.7)--(axis cs:26000,86.7);
  \node[font=\footnotesize,cMut,anchor=south west] at (axis cs:870,86.9)
     {internal ceiling (oracle best-of-$N$)};
  \addplot[cGnd,line width=1.4pt,mark=*,mark size=1.9,mark options={fill=cGnd,draw=white,line width=0.5pt}]
    coordinates {(1000,62.2)(5000,91.1)(20000,100.0)};
    \addlegendentry{proposer--reviewer \emph{(external)}}
  \addplot[cGnd2,line width=1.1pt,densely dashed,mark=square*,mark size=1.7,mark options={solid,fill=cGnd2,draw=white,line width=0.5pt}]
    coordinates {(1000,57.8)(5000,93.3)(20000,97.8)};
    \addlegendentry{single-agent loop \emph{(external)}}
  \addplot[cUng,line width=1.1pt,mark=triangle*,mark size=2.2,mark options={fill=cUng,draw=white,line width=0.5pt}]
    coordinates {(1000,80.0)(5000,77.8)(20000,86.7)};
    \addlegendentry{best-of-$N$, oracle verifier \emph{(internal)}}
  \addplot[cUng2,line width=1.1pt,densely dashed,mark=diamond*,mark size=2.2,mark options={solid,fill=cUng2,draw=white,line width=0.5pt}]
    coordinates {(1000,60.0)(5000,73.3)(20000,73.3)};
    \addlegendentry{reasoning-only \emph{(internal)}}
  \node[font=\scriptsize\bfseries,cGnd,anchor=south east,inner sep=1pt] at (axis cs:20000,100.4) {100.0};
  \node[font=\scriptsize,cUng2,anchor=north west,inner sep=2pt] at (axis cs:20000,73.0) {73.3};
\end{axis}
\end{tikzpicture}
\caption{\textbf{Internal scaling saturates; external scaling does not (Prediction 1).} Code pass rate vs.\ per-task token budget on the $15$ hard tasks ($3$-seed means; full table in \Cref{tab:scaling-curves}). Both internal strategies flatten as the budget grows: reasoning-only at its information ceiling, and best-of-$N$ at $86.7\%$ \emph{even with an oracle (ground-truth-test) verifier}, capped by the high-probability part of the model's own output distribution. The two external strategies import execution feedback each cycle and climb past that ceiling, the proposer--reviewer harness to a strict $100\%$ with zero seed variance. We run this experiment in \Cref{sec:interaction}.}
\label{fig:scaling}
\end{figure}

The argument is one sentence long: \emph{a model re-reading its own output cannot learn anything it did not already know.} A critique from the same weights that produced the artifact is post-processing, which adds no information about the correct answer; best-of-$N$ can only \emph{select} among the candidates the model actually draws, so even a perfect verifier cannot return a program it was never going to write. Longer thinking rearranges the known; more samples explore the already-likely. We call the resulting plateau the \textbf{internal ceiling} (formalized in \Cref{app:ceiling} via the data-processing inequality; since that bounds an information channel rather than achievable quality, our evidence is experimental). \Cref{fig:scaling} makes it concrete: reasoning-only and \emph{oracle-verified} best-of-$N$ flatten as the budget grows, while interaction strategies climb past them to a perfect pass rate. Grounded feedback escapes the ceiling because the observation is computed from the artifact's real behavior (a failing assertion, a measured bounding box, a search verdict); conditioning the next proposal on it lets the model synthesize a candidate it would never have sampled unaided. That is what separates interaction from selection (which cannot create a missing candidate) and from reasoning (which adds no information): a new information channel, not a constant-factor speedup.

\paragraph{The symmetric claim: an ungrounded metric cannot measure the gain.} The same argument applies to the \emph{scorer}: a model judging a lossy view of the artifact (a VLM reading a cropped screenshot) cannot certify an improvement that lives in the part its channel drops. Grounding the evaluation is therefore a precondition for \emph{detecting} interaction scaling wherever quality is not fully visible to the default judge, exactly the situation for rendered visual artifacts (\Cref{sec:grounded_eval}).

\subsection{Three testable predictions}
\label{sec:predictions}

\textbf{Prediction 1.} At a matched token budget, reasoning-only and best-of-$N$ scaling saturate strictly below what interaction reaches (\Cref{sec:interaction}). \textbf{Prediction 2.} Holding the task suite fixed and swapping only the reviewer's feedback channel, improvement tracks the instrument's \emph{coverage} of the defects present, not the act of reviewing (\Cref{sec:feedback}). \textbf{Prediction 3.} On modalities whose quality is invisible to the default judge, the gain is only detectable with a grounded metric, and an ungrounded reviewer can even regress quality (\Cref{sec:feedback,sec:grounded_eval}).

\section{Setup: A Budget-Aware Proposer--Reviewer Harness}
\label{sec:method}

\paragraph{Architecture.} The harness is pure scaffolding around a \emph{frozen} frontier model (Claude Sonnet~4, \texttt{claude-sonnet-4-20250514}, temperature~$0$ unless noted): no fine-tuning, no retrieval, no learned controller. A \textbf{proposer} produces an artifact. An \textbf{instrument} executes, renders, or measures it to produce a grounded signal. A \textbf{reviewer} (the same model in a diagnostic role) turns that signal into a structured list of defects, and the proposer revises. The loop repeats up to an iteration cap and \emph{keeps the best-scoring iteration}, so the reviewed result can never score below single-shot under the same metric. Single-shot is just one proposer call.

\paragraph{The three grounded instruments.} Every modality is paired with a deterministic or near-deterministic instrument that both \emph{feeds back} and \emph{scores}:
\begin{itemize}[leftmargin=1.4em,itemsep=1pt,topsep=2pt]
\item \textbf{Execution} (\texttt{pytest}): exact pass/fail plus the failing assertion and traceback. Each task ships a multi-assertion suite targeting the edge cases that separate correct from plausible-but-wrong (e.g.\ the semantic-version comparator is checked on pre-release precedence), and the deep-spec suite is additionally validated against a reference implementation. A ``pass'' therefore certifies the specified behavior, and the oracle best-of-$N$ baseline of \Cref{sec:interaction} selects against these \emph{same} tests, so its ceiling is measured by the identical criterion.
\item \textbf{Rendered geometry}: the artifact is rendered headless and every element's true bounding box is read from the layout engine. We compute, exactly: text-on-text overlap ($\ge 6$\,px), out-of-bounds/clipping, container overflow, document overflow ($>16$\,px), and \emph{box-group misalignment} (rows or columns of card-like boxes flagged for unequal size, misaligned far edges, or uneven gutters). Web pages are scored at desktop and mobile widths; animations are probed at each sampled frame.
\item \textbf{Native video} (Gemini~3.1~Pro): the rendered clip is judged whole against per-requirement binary checks, a near-grounded substitute where pixels, not a still, are observed.
\end{itemize}
Where no deterministic instrument exists (flowing web content, factual research), we fall back to a binary per-requirement rubric, flagged as such.

\begin{table}[t]
\centering
\small
\caption{Modality instantiations. Each modality pairs a grounded feedback channel (named per the taxonomy of \Cref{fig:taxonomy}) with the instrument that both drives revision and scores the result. ``$\dagger$'' marks the four modalities scored by the deterministic DOM-geometry instrument and analyzed jointly in \Cref{sec:grounded_eval}.}
\label{tab:setup}
\begin{tabular}{llll}
\toprule
Modality & Feedback channel & Grounded instrument & Hard suite ($N$) \\
\midrule
Code             & execution     & \texttt{pytest} pass/fail + traceback   & $15$ (+$11$ deep-spec) \\
Academic figures$^\dagger$ & geometry & DOM geometry + alignment           & $20$ \\
Slides$^\dagger$           & geometry & DOM geometry + alignment           & $12$ real-paper \\
Web pages$^\dagger$        & geometry & DOM geometry, multi-width          & $20$ \\
Animations$^\dagger$       & temporal & DOM geometry, per-frame            & $20$ SVG/CSS \\
Video editing    & exec.\ + temporal & script exec + Gemini-native rubric & $15$ \\
Deep research    & external facts & search-grounded factual rubric         & $15$ \\
\bottomrule
\end{tabular}
\end{table}

All artifacts are produced under a common \emph{design-principle} generation prompt (proximity, alignment, repetition, contrast, deliberate color) so that single-shot quality is already strong: the headroom we measure reflects genuine task difficulty, not a weak prompt.

\paragraph{Budget and protocol.} A run has a token budget $B$ split across proposer and reviewer calls; a sweep of the split (\Cref{tab:allocation}) shows pass rate rises monotonically with the proposer's share, so we allocate the majority to generation, and \Cref{tab:setup} lists the per-modality instruments. Unless noted, the proposer runs at temperature $0$, and each modality is evaluated over three independent seeds, keeping the best-scoring iteration (cap $\le 3$ for geometry, $\le 5$ for code); we compare single-shot against reviewed on each (task, seed) pair. For each result we report a two-sided paired sign test over the decisive pairs and, for the geometry and code effects, a paired-bootstrap $95\%$ confidence interval on the effect size; exact values accompany each figure.

\section{Interaction Scaling at a Matched Token Budget}
\label{sec:interaction}

\paragraph{The saturation experiment (Prediction 1).} On the hard code suite, at a fixed per-task token budget, we compare four strategies: reasoning-only (extended thinking), best-of-$N$ sampling with an \emph{oracle} verifier, a single-agent loop, and the proposer--reviewer harness. The result is \Cref{fig:scaling}, previewed in \Cref{sec:ceiling}: both internal strategies flatten as the budget grows, while \emph{every} strategy that iterates on feedback climbs past them toward a perfect pass rate. The best-of-$N$ comparison is the sharpest, because its verifier is the ground-truth test suite itself, so its ceiling is effectively pass@$N$: the tasks it never solves are those where none of its samples passes, and a perfect selector cannot return a candidate the proposer never writes. The harness solves those same tasks by conditioning the next draw on execution feedback. The gap is information-theoretic, not a tuning artifact.

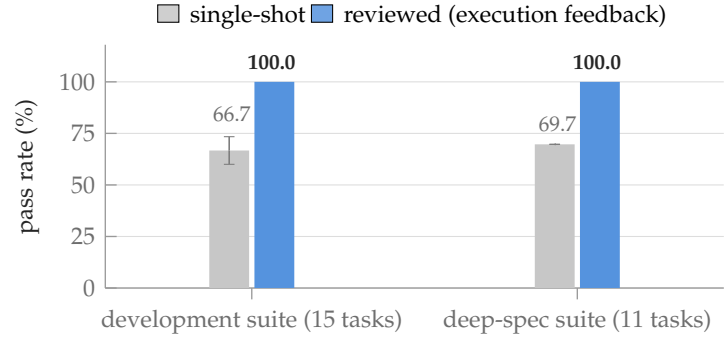
\begin{figure}[t]
\centering
\begin{tikzpicture}
\begin{axis}[paperbar,width=0.62\linewidth,height=4.8cm,
  ybar,bar width=15pt,ymin=0,ymax=118,
  ylabel={pass rate (\%)},
  symbolic x coords={dev,deepspec},xtick=data,
  xticklabels={development suite (15 tasks),deep-spec suite (11 tasks)},
  x tick label style={font=\footnotesize,color=cMut},
  enlarge x limits=0.45,ytick={0,25,50,75,100},ymajorgrids,
  legend style={at={(0.5,1.03)},anchor=south,legend columns=2,
    /tikz/every odd column/.append style={column sep=2pt}}]
  \addplot[swatch,draw=none,fill=black!22,
    error bars/.cd,y dir=both,y explicit,error bar style={cInk!60,line width=0.6pt}]
    coordinates {(dev,66.7) +- (0,6.7) (deepspec,69.7) +- (0,0)};
    \addlegendentry{single-shot}
  \addplot[swatch,draw=none,fill=cGnd!80]
    coordinates {(dev,100) (deepspec,100)};
    \addlegendentry{reviewed (execution feedback)}
  \node[font=\scriptsize\bfseries,cInk,anchor=south] at (axis cs:dev,101) {\hspace{15pt}100.0};
  \node[font=\scriptsize\bfseries,cInk,anchor=south] at (axis cs:deepspec,101) {\hspace{15pt}100.0};
  \node[font=\scriptsize,cMut,anchor=south] at (axis cs:dev,76) {\hspace{-15pt}66.7};
  \node[font=\scriptsize,cMut,anchor=south] at (axis cs:deepspec,71) {\hspace{-15pt}69.7};
\end{axis}
\end{tikzpicture}
\caption{\textbf{Execution feedback recovers every first-shot failure on both code suites.} Single-shot vs.\ harness-reviewed pass rate (development suite: $3$-seed mean, whisker $\pm1$\,SD; deep-spec: sign test $p=0.002$). Every failing run is recovered and no passing run regresses; the reviewed bars carry zero seed variance.}
\label{fig:code}
\end{figure}

\paragraph{Code as the clean case study.} Code is the cleanest demonstration, because the instrument is exact and the score is objective. The harness recovers every first-shot failure with no regressions, on both the development suite and a harder deep-spec suite of from-scratch implementations (a JSON parser, a spreadsheet-reference resolver, a minimal edit-script diff, an expression evaluator), lifting both to a perfect $100\%$ pass rate (\Cref{fig:code}). The mechanism is the same across recovered traces: turn~1 produces code that compiles but mishandles an edge case (an off-by-one boundary, semantic-version comparison, CIDR arithmetic); turn~2 receives the failing assertion, the offending input, and the expected-vs-actual values, and rewrites just that block; turn~3 confirms the fix. Reasoning alone cannot supply this: the model cannot know its boundary condition is wrong until a concrete test exercises it.

\begin{figure}[t]
\centering
\begin{tikzpicture}
\begin{axis}[paperbar,width=0.52\linewidth,height=3.6cm,
  xbar,xmin=0,xmax=1600,bar width=11pt,enlarge y limits=0.3,
  xlabel={mean output tokens per task at $B{=}20$K},
  symbolic y coords={loop,iad,harness},ytick={loop,iad,harness},
  yticklabels={single-agent loop,sample-and-select (IAD),proposer--reviewer},
  y tick label style={font=\footnotesize,color=cInk},
  xtick={0,500,1000,1500},xticklabels={0,500,1000,1500},xmajorgrids,clip=false]
  \addplot[draw=none,fill=cGnd!80] coordinates {(1029,harness)};
  \addplot[draw=none,fill=cGnd!45] coordinates {(1416,iad) (1431,loop)};
  \node[font=\scriptsize\bfseries,cInk,anchor=west,inner sep=2pt] at (axis cs:1029,harness) {1{,}029};
  \node[font=\scriptsize,cMut,anchor=west,inner sep=2pt,fill=white] at (axis cs:1416,iad) {1{,}416};
  \node[font=\scriptsize,cMut,anchor=west,inner sep=2pt,fill=white] at (axis cs:1431,loop) {1{,}431};
  \node[font=\scriptsize\bfseries,cGnd,anchor=west] at (axis cs:1680,harness) {100\% in 3/3 seeds};
  \node[font=\scriptsize,cMut,anchor=west] at (axis cs:1680,iad) {100\% (1 seed)};
  \node[font=\scriptsize,cMut,anchor=west] at (axis cs:1680,loop) {100\% in 2/3 seeds};
\end{axis}
\end{tikzpicture}
\caption{\textbf{Same ceiling, different cost and reliability.} All three interaction strategies use execution feedback and tie on pass rate within seed noise (sign test $p>0.6$); the proposer--reviewer harness converges with ${\sim}28\%$ fewer tokens and is the only variant at $100\%$ across all seeds (\Cref{tab:LvsH}).}
\label{fig:efficiency}
\end{figure}
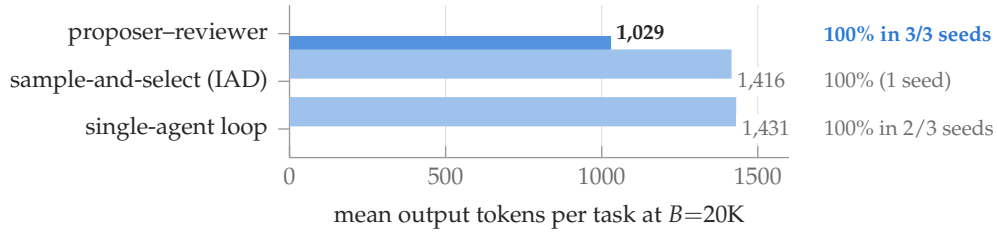

\paragraph{Architecture buys efficiency and reliability, not ceiling.} Among interaction strategies, the ceiling pass rate ties within seed noise; what separates them is cost and variance (\Cref{fig:efficiency}). The proposer--reviewer harness is the most token-efficient, and the only variant that reaches $100\%$ in \emph{every} seed. Three things explain the efficiency: the reviewer sees only what it needs to locate defects, it returns a structured list rather than raw \texttt{stderr}, and the two roles are specialized.

\paragraph{Built-in early stopping.} The loop does not over-revise: on code, \emph{every} already-passing single-shot task is submitted at iteration~1 with no revision and no wasted tokens. The grounded signal (a failing test) is the trigger; without it the loop is silent. This is why the harness's token overhead over single-shot stays modest despite a generous iteration cap.

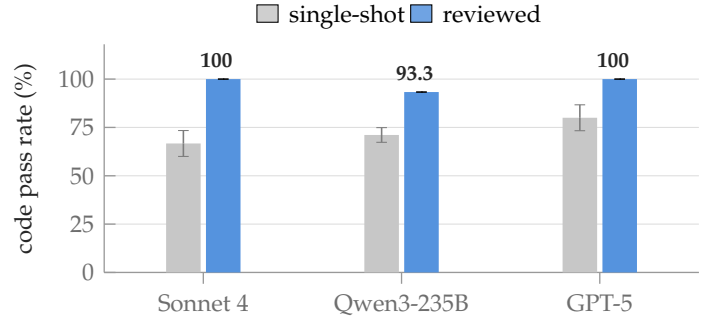
\begin{figure}[t]
\centering
\begin{tikzpicture}
\begin{axis}[paperbar,width=0.6\linewidth,height=4.6cm,
  ybar,bar width=13pt,ymin=0,ymax=118,
  ylabel={code pass rate (\%)},
  symbolic x coords={Sonnet 4,Qwen3-235B,GPT-5},xtick=data,
  x tick label style={font=\footnotesize,color=cMut},
  enlarge x limits=0.25,ymajorgrids,ytick={0,25,50,75,100},
  legend style={at={(0.5,1.03)},anchor=south,legend columns=2,
    /tikz/every odd column/.append style={column sep=2pt}}]
  \addplot[swatch,draw=none,fill=black!22,
    error bars/.cd,y dir=both,y explicit,error bar style={cInk!60,line width=0.6pt}]
    coordinates {(Sonnet 4,66.7) +- (0,6.7) (Qwen3-235B,71.1) +- (0,3.8) (GPT-5,80.0) +- (0,6.7)};
    \addlegendentry{single-shot}
  \addplot[swatch,draw=none,fill=cGnd!80,
    error bars/.cd,y dir=both,y explicit]
    coordinates {(Sonnet 4,100) +- (0,0) (Qwen3-235B,93.3) +- (0,0) (GPT-5,100) +- (0,0)};
    \addlegendentry{reviewed}
  \node[font=\scriptsize\bfseries,cInk,anchor=south] at (axis cs:Sonnet 4,101) {\hspace{10pt}100};
  \node[font=\scriptsize\bfseries,cInk,anchor=south] at (axis cs:Qwen3-235B,94.3) {\hspace{10pt}93.3};
  \node[font=\scriptsize\bfseries,cInk,anchor=south] at (axis cs:GPT-5,101) {\hspace{10pt}100};
\end{axis}
\end{tikzpicture}
\caption{\textbf{The execution loop replicates across model families.} Single-shot vs.\ harness-reviewed code pass rate for three families, three seeds each (lifts $+33.3$/$+22.2$/$+20.0$\,pp; whiskers $\pm1$\,SD across seeds). The reviewed bars carry \emph{zero} seed variance for all three families.}
\label{fig:crossmodel-code}
\end{figure}

\paragraph{The effect is architectural, not Claude-specific.} Over three seeds per family, the harness lifts all three we test: Sonnet~4, Qwen3-235B, and GPT-5 (the last from a higher single-shot baseline; \Cref{fig:crossmodel-code}). The telling detail is the variance: single-shot fluctuates seed-to-seed for every family, while the reviewed ceiling has \emph{zero} seed variance for all three, converging to the same point regardless of the starting draw. The result also holds out of sample: on a $32$-task held-out suite built after the method was fixed, the harness again recovers every first-shot failure with no regressions and reaches a perfect pass rate (\Cref{tab:heldout}; the absolute lift is smaller only because the held-out single-shot baseline is higher).

\section{Feedback-Side Grounding: Swapping the Reviewer's Instrument}
\label{sec:feedback}

The saturation experiment compared strategies. This section holds everything else fixed (task suite, proposer model, number of reviewing passes) and swaps \emph{only the reviewer's feedback channel}, testing the coverage principle (Prediction 2) directly on both a textual and a visual modality.

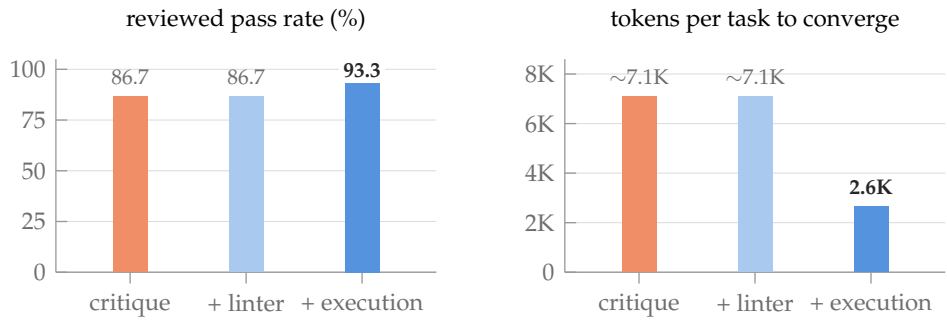
\begin{figure}[t]
\centering
\begin{tikzpicture}
\begin{axis}[paperbar,name=axpass,width=0.42\linewidth,height=4.4cm,
  ybar,bar width=13pt,ymin=0,ymax=105,
  title={\footnotesize reviewed pass rate (\%)},
  symbolic x coords={critique,linter,execution},xtick={critique,linter,execution},
  xticklabels={critique,{+ linter},{+ execution}},
  x tick label style={font=\footnotesize,color=cMut},
  enlarge x limits=0.32,ytick={0,25,50,75,100},ymajorgrids]
  \addplot[draw=none,fill=cUng!75,bar shift=0pt] coordinates {(critique,86.7)};
  \addplot[draw=none,fill=cGnd!40,bar shift=0pt] coordinates {(linter,86.7)};
  \addplot[draw=none,fill=cGnd!80,bar shift=0pt] coordinates {(execution,93.3)};
  \node[font=\scriptsize,cMut,anchor=south] at (axis cs:critique,87.5) {86.7};
  \node[font=\scriptsize,cMut,anchor=south] at (axis cs:linter,87.5) {86.7};
  \node[font=\scriptsize\bfseries,cInk,anchor=south,fill=white,inner sep=1.5pt] at (axis cs:execution,94) {93.3};
\end{axis}
\begin{axis}[paperbar,at={($(axpass.east)+(1.7cm,0)$)},anchor=west,
  width=0.42\linewidth,height=4.4cm,
  ybar,bar width=13pt,ymin=0,ymax=8600,
  title={\footnotesize tokens per task to converge},
  symbolic x coords={critique,linter,execution},xtick={critique,linter,execution},
  xticklabels={critique,{+ linter},{+ execution}},
  x tick label style={font=\footnotesize,color=cMut},
  enlarge x limits=0.32,ytick={0,2000,4000,6000,8000},
  yticklabels={0,2K,4K,6K,8K},ymajorgrids]
  \addplot[draw=none,fill=cUng!75,bar shift=0pt] coordinates {(critique,7100)};
  \addplot[draw=none,fill=cGnd!40,bar shift=0pt] coordinates {(linter,7100)};
  \addplot[draw=none,fill=cGnd!80,bar shift=0pt] coordinates {(execution,2643)};
  \node[font=\scriptsize,cMut,anchor=south] at (axis cs:critique,7200) {$\sim$7.1K};
  \node[font=\scriptsize,cMut,anchor=south] at (axis cs:linter,7200) {$\sim$7.1K};
  \node[font=\scriptsize\bfseries,cInk,anchor=south] at (axis cs:execution,2750) {2.6K};
\end{axis}
\end{tikzpicture}
\caption{\textbf{Swapping only the feedback signal on a fixed code suite: improvement tracks coverage, and cost tracks it decisively.} One reviewing pass per configuration. Ungrounded critique (orange), critique + linter (light blue: grounded, but observing only \emph{form}), and critique + test execution (blue: grounded, observing \emph{behavior}). The linter configuration matches bare critique on these runtime-logic bugs (exactly what the coverage principle predicts) while the execution configuration reaches a higher ceiling at ${\sim}2.5\times$ lower cost and fewer iterations ($1.53$ vs.\ $2.07$; per-seed detail in \Cref{tab:type-controls}).}
\label{fig:typecontrol}
\end{figure}

\paragraph{On code: execution feedback wins where its coverage lies.} We run three configurations on the fixed code suite, each with \emph{exactly one} reviewing pass: ungrounded critique (the LLM re-reads the code), static form (the same critique plus \texttt{ruff} linter output), and execution (the same critique plus the failing test's traceback). Any difference between them therefore isolates the \emph{signal}, not the act of reviewing. \Cref{fig:typecontrol} shows both outcomes. The gap in pass rate is modest and rests on a single leaner-budget seed, so we read it cautiously. The clear gap is in \emph{cost}: the execution configuration converges roughly $2.5\times$ cheaper and in fewer iterations, because a concrete failing assertion pinpoints the fix where an ungrounded critique can only guess. The static-form configuration is the coverage principle's cleanest test: the linter is genuinely grounded, but the seeded bugs are runtime-logic errors that leave no trace in the source, so it buys nothing over bare critique. Grounding helps only as far as the instrument can see.\footnote{Symmetrically, the principle predicts a static-form lift on a bug set that \emph{is} visible in form (type confusions catchable by \texttt{mypy}, taint patterns catchable by \texttt{semgrep}). Constructing that suite is future work; on our behavioral bug set the null result is the prediction.}

\begin{figure}[t]
\centering
\begin{tikzpicture}[
  dumb/.style={line width=1.6pt,-{Latex[length=2.6mm,width=2.2mm]}},
  ssdot/.style={circle,draw=cMut,fill=white,line width=1pt,inner sep=0pt,minimum size=7pt}]
\begin{axis}[name=axslide,width=0.44\linewidth,height=5.0cm,
  title={\footnotesize dense slides},
  ylabel={mean geometric defects per artifact},
  xmin=0.35,xmax=2.65,ymin=0,ymax=2.9,ytick={0,1,2},
  xtick={1,2},xticklabels={VLM\\reviewer,geometric\\reviewer},
  x tick label style={font=\footnotesize,color=cMut,align=center},
  ymajorgrids,axis x line*=bottom,axis y line*=left]
  \draw[dumb,cUng] (axis cs:1,1.89) -- (axis cs:1,2.44);
  \node[ssdot] at (axis cs:1,1.89) {};
  \node[font=\scriptsize,cMut,anchor=east,inner sep=6pt] at (axis cs:1,1.89) {1.89};
  \node[font=\scriptsize\bfseries,cUng!85!black,anchor=west,inner sep=4pt] at (axis cs:1,2.44) {2.44 \ worse};
  \draw[dumb,cGnd] (axis cs:2,1.25) -- (axis cs:2,0.33);
  \node[ssdot] at (axis cs:2,1.25) {};
  \node[font=\scriptsize,cMut,anchor=west,inner sep=6pt] at (axis cs:2,1.25) {1.25};
  \node[font=\scriptsize\bfseries,cGnd,anchor=east,inner sep=4pt] at (axis cs:2,0.33) {0.33 \ fixed};
\end{axis}
\begin{axis}[at={($(axslide.east)+(1.5cm,0)$)},anchor=west,
  width=0.44\linewidth,height=5.0cm,
  title={\footnotesize academic figures},
  xmin=0.35,xmax=2.65,ymin=0,ymax=0.8,ytick={0,0.25,0.5,0.75},
  xtick={1,2},xticklabels={VLM\\reviewer,geometric\\reviewer},
  x tick label style={font=\footnotesize,color=cMut,align=center},
  ymajorgrids,axis x line*=bottom,axis y line*=left]
  \draw[dumb,cUng] (axis cs:1,0.52) -- (axis cs:1,0.62);
  \node[ssdot] at (axis cs:1,0.52) {};
  \node[font=\scriptsize,cMut,anchor=east,inner sep=6pt] at (axis cs:1,0.52) {0.52};
  \node[font=\scriptsize,cUng!85!black,anchor=south,inner sep=3pt] at (axis cs:1,0.64) {0.62 \ worse (n.s.)};
  \draw[dumb,cGnd] (axis cs:2,0.57) -- (axis cs:2,0.15);
  \node[ssdot] at (axis cs:2,0.57) {};
  \node[font=\scriptsize,cMut,anchor=west,inner sep=6pt] at (axis cs:2,0.57) {0.57};
  \node[font=\scriptsize\bfseries,cGnd,anchor=east,inner sep=4pt] at (axis cs:2,0.15) {0.15 \ fixed};
\end{axis}
\end{tikzpicture}
\caption{\textbf{Same tasks, same proposer, one reviewing pass: the arrow's direction is set by the reviewer's instrument.} Each arrow runs from that configuration's own single-shot baseline (open circle; the configurations are separate runs, hence differing baselines) to its reviewed result, in mean geometric defects per artifact (lower is better; note the differing $y$ scales). The screenshot-fed VLM reviewer moves \emph{up} on slides and directionally up on figures; the deterministic geometric reviewer moves sharply \emph{down} on both (slides $-73\%$, $p=0.0018$; figures $-74\%$, $p=7{\times}10^{-4}$).}
\label{fig:ablation}
\end{figure}
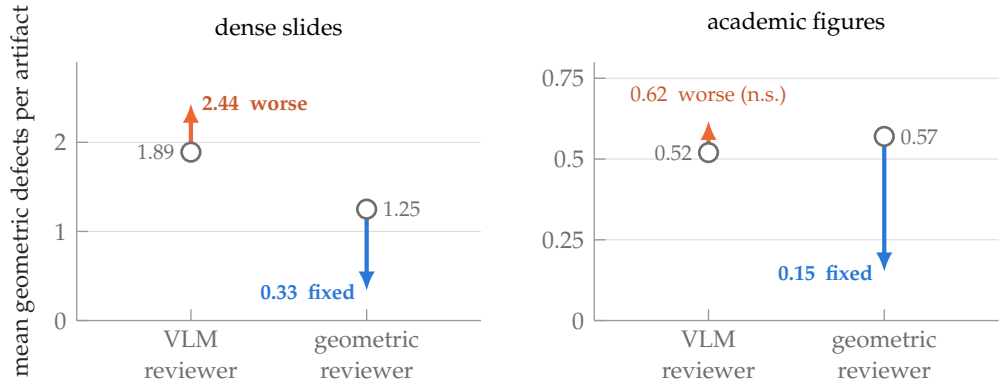

\paragraph{On slides and figures: an uncovered instrument makes things worse.} The decisive version of this control is visual. We fix the task suite and the proposer, and give the reviewer either (i) the deterministic geometry report (measured boxes, exact overlaps) or (ii) a VLM's reading of a screenshot, which crops the very defects at issue. The two configurations are separate runs, so each is compared against its \emph{own} single-shot baseline; what matters is the \emph{direction} each reviewer moves that baseline (\Cref{fig:ablation}). The geometry reviewer sharply reduces real layout defects on both modalities. The VLM reviewer \emph{increases} defects on slides and is directionally worse on figures: blind to the true geometry, its edits break alignment about as often as they fix it. One reviewing pass, opposite sign, decided entirely by whether the signal covers the defect. (On the same slides, the binary VLM \emph{rubric} saturates at the same time; \Cref{sec:grounded_eval} takes up that evaluation-side half.)

\paragraph{What this establishes.} On the feedback side, grounding is necessary, but the quantity that decides the outcome is \emph{coverage}. A grounded instrument that cannot see the defect (the linter, on runtime bugs) matches ungrounded critique, and an instrument that actively drops the defect (the screenshot) is worse than no reviewer at all. The remaining question is symmetric: what happens when the \emph{metric} is such an instrument?

\section{Evaluation-Side Grounding: Deterministic Instruments vs.\ Model Judges}
\label{sec:grounded_eval}

The visual modalities expose the symmetric half of the framework (Prediction 3): even when grounded feedback \emph{does} fix the artifact, an ungrounded metric cannot see the fix.

\begin{figure}[t]
\centering
\begin{subfigure}[c]{0.35\linewidth}
\centering
\begin{tikzpicture}
\begin{axis}[paperbar,width=\linewidth,height=4.4cm,
  ybar,bar width=20pt,ymin=0,ymax=16,
  ylabel={\footnotesize figures rated clean (of 15)},
  symbolic x coords={VLM judge,DOM geometry},xtick={VLM judge,DOM geometry},
  xticklabels={VLM\\judge,DOM\\geometry},
  x tick label style={font=\footnotesize,color=cMut,align=center},
  enlarge x limits=0.55,ymajorgrids,ytick={0,5,10,15}]
  \addplot[draw=none,fill=cUng!75,bar shift=0pt] coordinates {(VLM judge,14)};
  \addplot[draw=none,fill=cGnd!80,bar shift=0pt] coordinates {(DOM geometry,3)};
  \node[font=\footnotesize\bfseries,cInk,anchor=south,fill=white,inner sep=1.5pt] at (axis cs:VLM judge,14.2) {14};
  \node[font=\footnotesize\bfseries,cInk,anchor=south] at (axis cs:DOM geometry,3.2) {3};
\end{axis}
\end{tikzpicture}
\caption{The VLM passes $14/15$; the deterministic instrument, $3/15$.}
\label{fig:blind}
\end{subfigure}\hfill
\begin{subfigure}[c]{0.62\linewidth}
\centering
\includegraphics[width=\linewidth]{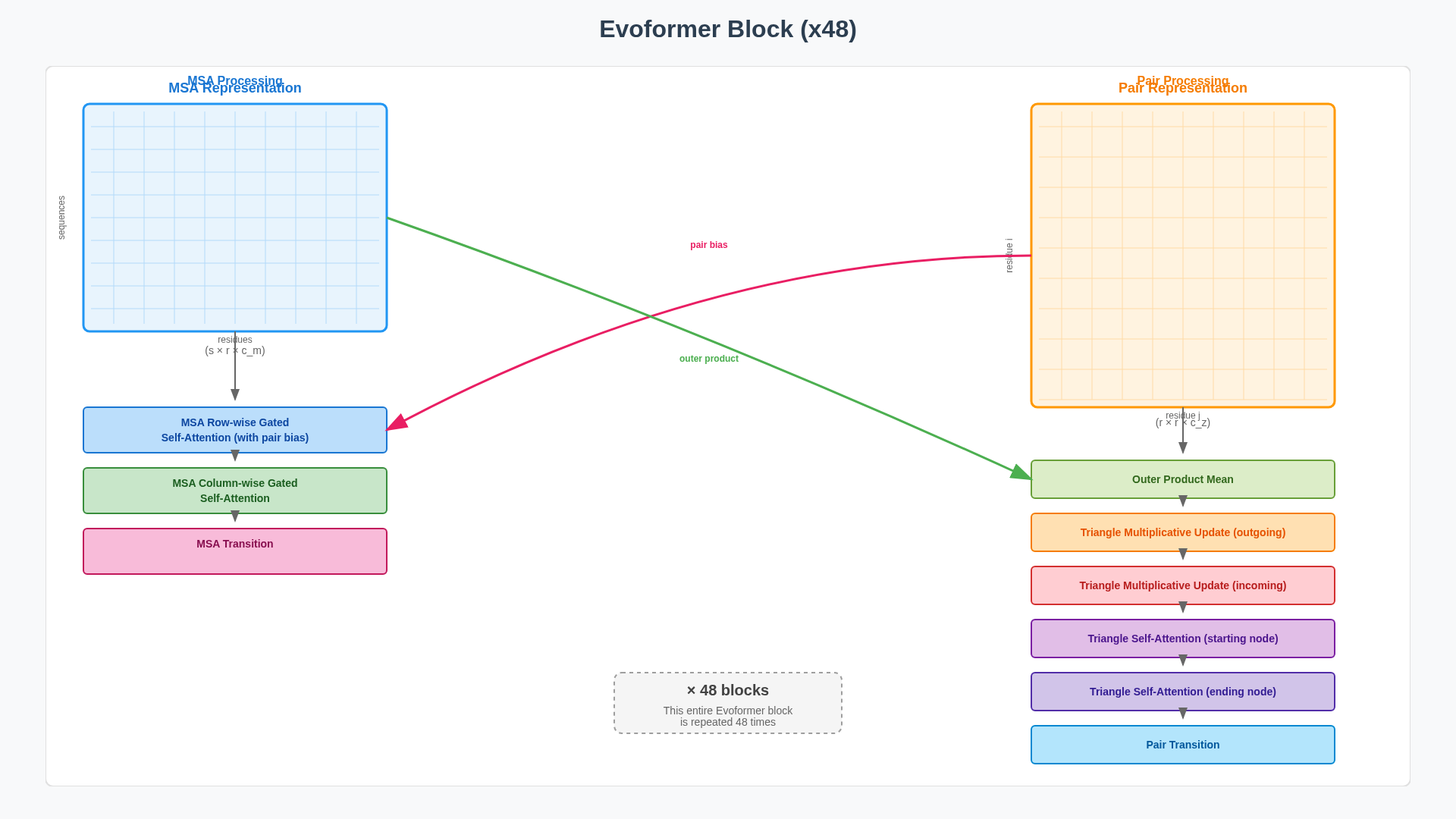}
\caption{A single-shot figure the VLM rated ``perfect'': both section titles are superimposed (``MSA Processing'' over ``MSA Representation''; ``Pair Processing'' over ``Pair Representation'') and axis labels collide with the matrix borders.}
\label{fig:evoformer}
\end{subfigure}
\caption{\textbf{The default VLM judge is structurally blind to layout defects.} \textbf{(\subref{fig:blind})} On the same $15$ single-shot academic figures, the VLM-on-a-screenshot judge rates $14$ ``perfect'' while the DOM-geometry instrument finds only $3$ actually clean; the other $11$ are broken in ways cropped off-frame or below VLM acuity. \textbf{(\subref{fig:evoformer})} A representative ``perfect''-rated render whose text-on-text overlaps are exactly what the geometry check flags and the screenshot judge misses.}
\label{fig:blindpair}
\end{figure}
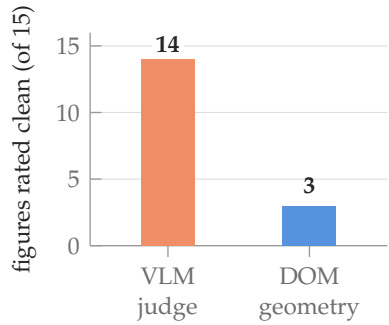
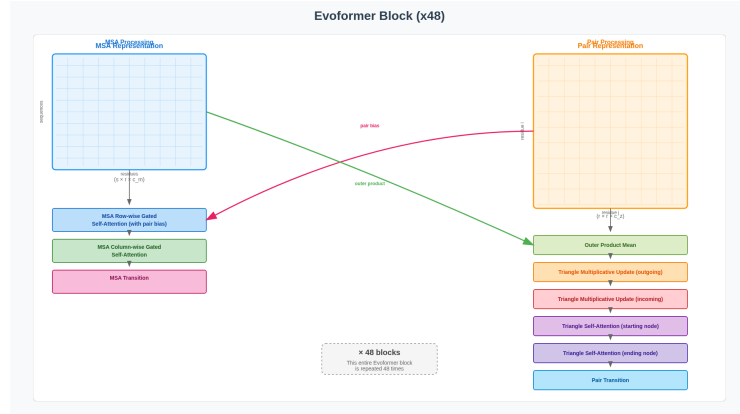

\paragraph{The default VLM judge is structurally blind.} The field's default evaluator for visual artifacts is a binary per-requirement rubric scored by a VLM from a screenshot. On an initial probe of $15$ dense academic-paper architecture figures (even with native-resolution quadrant tiling), this judge rates $14$ of $15$ single-shot renders ``perfect,'' meaning they satisfy every \emph{content} requirement it checks. Yet by direct inspection most of those renders are broken, with section titles printed on top of each other and labels spilling out of boxes (\Cref{fig:evoformer}). The failure is mechanical, and no better prompt fixes it: screenshots are captured at a fixed resolution, so content overflowing the canvas is cropped \emph{off-frame before the image reaches the judge}, and small in-frame overlaps fall below its acuity. The DOM-geometry instrument, reading the same artifacts' actual bounding boxes, finds only $3$ of $15$ truly clean (\Cref{fig:blind}). This is not a wrong answer to the same question; the judge is \emph{structurally unable to observe} the property that matters, an instrument whose coverage excludes the defect, used as the score. (This probe is a deliberately hard, unconstrained set; the suites scored below use the design-principle prompt of \Cref{sec:method} and are correspondingly cleaner single-shot.)

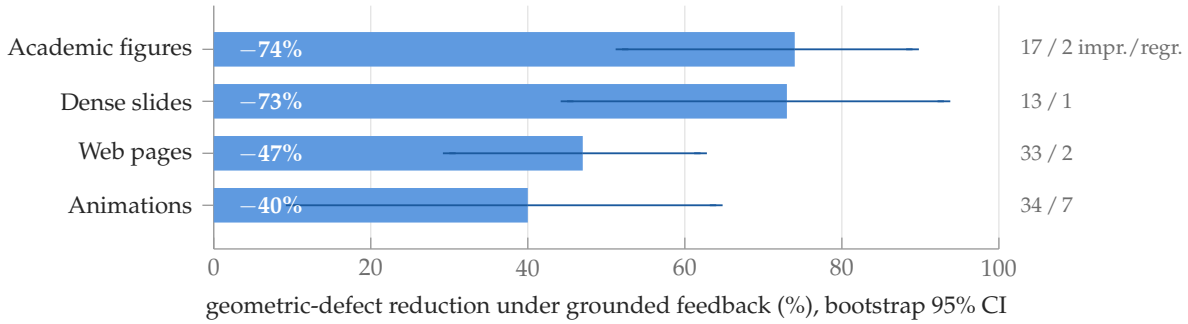
\begin{figure}[t]
\centering
\begin{tikzpicture}
\begin{axis}[paperbar,width=0.76\linewidth,height=4.8cm,
  xbar,xmin=0,xmax=100,bar width=13pt,enlarge y limits=0.28,clip=false,
  xlabel={geometric-defect reduction under grounded feedback (\%), bootstrap 95\% CI},
  symbolic y coords={Animations,Web pages,Dense slides,Academic figures},
  ytick=data,y tick label style={font=\footnotesize,color=cInk},
  xtick={0,20,40,60,80,100},xmajorgrids]
  \addplot[draw=none,fill=cGnd!75,
    error bars/.cd,x dir=both,x explicit,
    error bar style={cGndDk,line width=0.7pt},
    error mark options={cGndDk,mark size=2.4pt,line width=0.7pt}]
    coordinates {(74,Academic figures)+=(15,0)-=(22,0)
                 (73,Dense slides)+=(20,0)-=(28,0)
                 (47,Web pages)+=(15,0)-=(17,0)
                 (40,Animations)+=(24,0)-=(30,0)};
  \node[font=\footnotesize\bfseries,white,anchor=west] at (axis cs:1.8,Academic figures) {$-$74\%};
  \node[font=\footnotesize\bfseries,white,anchor=west] at (axis cs:1.8,Dense slides) {$-$73\%};
  \node[font=\footnotesize\bfseries,white,anchor=west] at (axis cs:1.8,Web pages) {$-$47\%};
  \node[font=\footnotesize\bfseries,white,anchor=west] at (axis cs:1.8,Animations) {$-$40\%};
  \node[font=\scriptsize,cMut,anchor=west] at (axis cs:101.5,Academic figures) {17\,/\,2 impr./regr.};
  \node[font=\scriptsize,cMut,anchor=west] at (axis cs:101.5,Dense slides) {13\,/\,1};
  \node[font=\scriptsize,cMut,anchor=west] at (axis cs:101.5,Web pages) {33\,/\,2};
  \node[font=\scriptsize,cMut,anchor=west] at (axis cs:101.5,Animations) {34\,/\,7};
\end{axis}
\end{tikzpicture}
\caption{\textbf{Grounded geometric feedback removes real layout defects on all four visual modalities.} Mean defect reduction under one identical configuration ($3$ seeds; whiskers are paired-bootstrap 95\% CIs, all excluding zero; right column counts improved vs.\ regressed task-runs; all paired sign tests $p<2{\times}10^{-3}$; full statistics in \Cref{tab:geometric}). The standard VLM-on-a-screenshot metric measures none of this.}
\label{fig:geomreduction}
\end{figure}

\paragraph{A grounded instrument reveals a large, decisive effect on all four modalities.} We score four visual modalities with the same DOM-geometry-plus-alignment instrument under one identical configuration (Sonnet~4, temperature~$0$, design-principle prompt, three seeds, propose $\to$ measure $\to$ feed exact defects back $\to$ revise, $\le3$ iterations). \Cref{fig:geomreduction} summarizes and \Cref{tab:geometric} gives the full statistics: grounded geometry feedback removes a large fraction of the real layout defects on \emph{all four} modalities, with every bootstrap confidence interval excluding zero and improvements outnumbering regressions by roughly ten to one. (\Cref{fig:beforeafter} in the introduction shows the per-artifact reality behind these aggregates.) The size of the effect tracks single-shot headroom: dense responsive web pages are full of defects out of the box, while a frontier model under a design-principle prompt already lays out many figures and slides cleanly. Animations carry the widest interval, because their per-task defect counts are heavy-tailed; there we treat the sign test as the robust statement and the mean as indicative.

\paragraph{Is the reduction circular?} The instrument both \emph{drives} the revision and \emph{scores} it, and the harness keeps the best-scoring iteration, so \emph{some} reduction is mechanically guaranteed. The real questions are how large it is, and whether it reflects quality a human would care about. Three facts argue it is not just an artifact of optimizing the reported number. \textbf{(i)~The defects are real:} single-shot renders carry them at high rate (\Cref{fig:evoformer}), and the thresholds are conservative, so a flagged defect is visible, not sub-perceptual. \textbf{(ii)~The magnitude is not preordained:} keeping the best of three iterations could move the metric by almost nothing; that grounded revision instead removes most of the real defects (\Cref{fig:geomreduction}) is a property of the feedback, not of the keep-best rule. \textbf{(iii)~The ungrounded-reviewer ablation is the decisive control:} scored by the \emph{same} metric, the VLM-feedback configuration fails to reduce it and worsens slides (\Cref{fig:ablation}). So the reduction tracks the grounding of the feedback, not the act of optimizing the scorer, a conclusion the model-free cross-model replication below reinforces. The principal remaining caveat is a quantitative human-preference study confirming that DOM-defect reduction maps onto perceived quality across the full suite.

\paragraph{The geometry effect is not proposer-specific.} To rule out a Claude- or prompt-specific explanation, we run the same geometry-feedback harness with \textbf{Gemini~3.1~Pro} as proposer on the dense real-paper slide suite, scored by the identical model-free DOM instrument. The effect reproduces and is larger: Gemini's single-shot slides carry more defects than Sonnet's, and the grounded loop removes almost all of the excess ($-93\%$ vs.\ $-73\%$ for Sonnet, $19$ of $20$ decisive task-runs improved). Since the instrument uses no model, the effect is neither proposer- nor scorer-specific.

\paragraph{Two modalities where the honest answer is ``no headroom.''} \emph{Video editing}, scored by Gemini~3.1~Pro's native full-clip rubric (near-grounded, since it observes the pixels, not a still), is already strong single-shot, so the reviewed lift is small and not significant; a hardened multi-step suite de-saturates it. \emph{Deep research}, scored against exact provided facts, saturates: a frontier model knows well-documented facts and does not fall for the planted traps, so the factual-feedback lift is capped by a near-zero single-shot error rate. This is a genuine limit of the modality, since de-saturating it would require facts the judge itself cannot grade. We report both as scoped negatives, not headline lifts (\Cref{fig:headline}).

\section{Internalizing the Harness into a Small Student}
\label{sec:distill}

Two things could carry the harness's gains: the grounded scaffolding around the model, or behavior absorbed into the weights. This section asks how much is the latter: whether the loop's interaction quality can be \emph{internalized}, so a small model produces high-quality artifacts in fewer passes. This complements the thesis rather than replacing it, because the cheaper internalized proposer still runs \emph{inside} the same grounded harness. We distill judge-filtered teacher trajectories from the harness into an $8$B Qwen3-VL student by supervised fine-tuning.

\begin{figure}[t]
\centering
\begin{tikzpicture}
\begin{axis}[paperbar,width=0.6\linewidth,height=4.6cm,clip=false,
  ybar,bar width=26pt,ymin=0,ymax=1.2,
  ylabel={capability vs.\ teacher ($\times$)},
  symbolic x coords={mean@1,pass@2},xtick=data,enlarge x limits=0.6,
  x tick label style={font=\footnotesize,color=cMut},
  ymajorgrids,ytick={0,0.25,0.5,0.75,1.0},yticklabel={\pgfmathprintnumber{\tick}$\times$}]
  \fill[black!5] ({rel axis cs:0,0}|-{axis cs:mean@1,1.0}) rectangle ({rel axis cs:1,0}|-{axis cs:mean@1,1.2});
  \draw[cMut,densely dashed,line width=0.7pt] ({rel axis cs:0,0}|-{axis cs:mean@1,1.0}) -- ({rel axis cs:1,0}|-{axis cs:mean@1,1.0});
  \node[font=\footnotesize,cMut,anchor=south east] at ({rel axis cs:0.99,0}|-{axis cs:mean@1,1.01}) {teacher ($1\times$)};
  \addplot[draw=none,fill=cStu!65] coordinates {(mean@1,0.51) (pass@2,0.70)};
  \node[font=\footnotesize\bfseries,cInk,anchor=south] at (axis cs:mean@1,0.52) {0.51$\times$};
  \node[font=\footnotesize\bfseries,cInk,anchor=south] at (axis cs:pass@2,0.71) {0.70$\times$};
  \node[font=\scriptsize,cGnd!80!black,anchor=center]
    at ($(axis cs:mean@1,0.90)!0.5!(axis cs:pass@2,0.90)$)
    {sampling adds $+0.19\times$};
\end{axis}
\end{tikzpicture}
\caption{\textbf{An $8$B student internalizes much of the teacher's interaction quality.} Distilled student as a fraction of the frontier teacher on the $44$-task out-of-distribution suite (\Cref{tab:distill-ood}): one sample recovers $0.51\times$ and two samples $0.70\times$ of teacher capability, at ${\sim}10\times$ lower deployment cost. On the $18$-task hard held-out suite the same student passes $44\%$ at pass@$1$ and $56\%$ at pass@$3$ (\Cref{tab:distill-headline}).}
\label{fig:distill}
\end{figure}
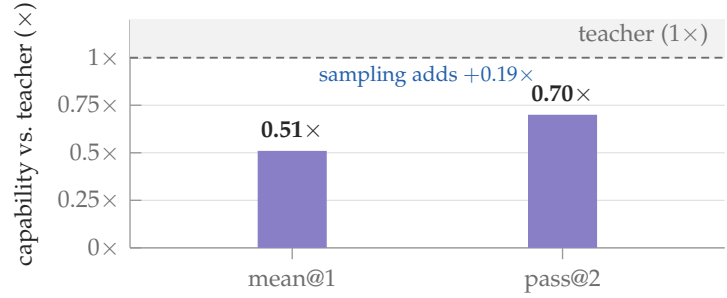

\paragraph{Result.} On an out-of-distribution suite, the student recovers about half the teacher's single-sample capability, and a second sample closes much of the remaining gap (\Cref{fig:distill}), at roughly $10\times$ lower deployment cost. On a harder held-out suite of the teacher's own pre-curated failures it passes a substantial fraction (\Cref{tab:distill-headline}), and run back inside the harness it recovers still more.

\begin{figure}[t]
\centering
\begin{tikzpicture}
\begin{axis}[width=0.56\linewidth,height=4.6cm,
  xlabel={samples $k$},ylabel={held-out judge-keep (\%)},
  xmin=0.8,xmax=3.6,ymin=20,ymax=62,clip=false,
  xtick={1,2,3},xticklabels={mean@1,pass@2,pass@3},
  x tick label style={font=\footnotesize,color=cMut},
  ytick={25,35,45,55},ymajorgrids,
  axis x line*=bottom,axis y line*=left]
  \addplot[cStu,line width=1.4pt,mark=*,mark size=1.9,
    mark options={fill=cStu,draw=white,line width=0.5pt}]
    coordinates {(1,31)(2,50)(3,56)};
  \addplot[cUng,line width=1.1pt,mark=square*,mark size=1.7,
    mark options={fill=cUng,draw=white,line width=0.5pt}]
    coordinates {(1,28)(2,39)(3,39)};
  \node[font=\scriptsize\bfseries,cStu,anchor=west,inner sep=3pt] at (axis cs:3,56) {SFT student: 56};
  \node[font=\scriptsize,cUng!85!black,anchor=west,inner sep=3pt] at (axis cs:3,39) {+\,RFT: 39};
\end{axis}
\end{tikzpicture}
\caption{\textbf{RFT polishes consistency but spends the variance that sampling needs.} Judge-keep vs.\ number of samples on the $18$-task hard held-out suite for the chosen SFT student and the same student after RFT. RFT improves every per-turn consistency metric (\Cref{tab:distill-rft}) yet regresses pass@$3$ by $17$\,pp: the near-tie at one sample and the flat curve past it show the lost quality is exactly the distributional spread best-of-$N$ converts into solved tasks.}
\label{fig:variance}
\end{figure}
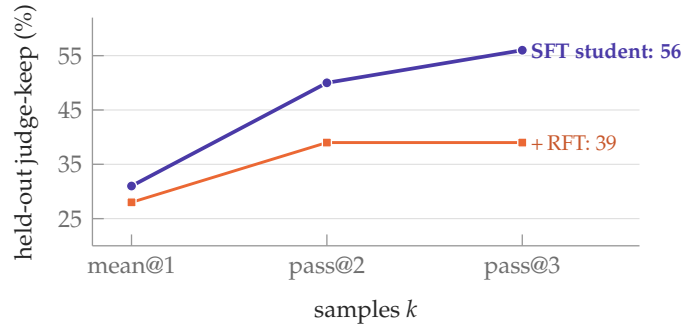

\paragraph{Variance is a budget (a cautionary finding).} Reinforcement fine-tuning (RFT) on top of SFT improves every per-turn consistency metric we measure, \emph{and} it lowers pass@$k$ (\Cref{fig:variance}). The two students start nearly tied on a single sample; as samples are added, the SFT student keeps turning fresh draws into newly solved tasks, while the RFT student's curve goes flat. When you deploy by sampling, the variance in the output distribution \emph{is} the lever that inference scaling pulls, so post-training that reduces that variance spends from the very budget best-of-$N$ relies on. Variance here is not noise to minimize; it is the resource sampling converts into quality. The recipe, then: use SFT to internalize interaction quality, keep the sampling temperature, and run the student inside the grounded harness.

\section{Related Work}
\label{sec:related}

\paragraph{Test-time compute scaling.} Two axes are established. \emph{Reasoning} scaling (chain-of-thought, tree search, and extended thinking \citep{wei2022chain,yao2023tree,deepseek2025r1,snell2024scaling,kimi2025k15}) spends more tokens before committing. \emph{Sampling} scaling (self-consistency and best-of-$N$ \citep{wang2023selfconsistency,brown2024large}) draws multiple attempts and selects one. Both only rework the same frozen weights and prompt (both are \emph{internal} in the sense of \Cref{sec:framework}, which makes precise how they are bounded). We position \emph{interaction} as a third, external axis that imports information the model does not have.

\paragraph{Self-correction and critique.} Iterative refinement with model-generated feedback (Reflexion \citep{shinn2023reflexion}, ReAct \citep{yao2023react}, Self-Refine \citep{madaan2023selfrefine}, CRITIC \citep{gou2024critic}) is widely used, but its reliability is contested: \citet{huang2024selfcorrect} show that models often cannot self-correct reasoning without an external signal, and \citet{kamoi2025correct} chart when correction does and does not help. Our taxonomy makes the distinction explicit: \emph{ungrounded} critique is subject to the internal ceiling, while refinement against an executed or measured observation is not. Our controls (\Cref{sec:feedback}) then show that it is the instrument's coverage of the defect, not the extra critique pass, that carries the gain.

\paragraph{Grounded and execution feedback.} Tool use and execution feedback (Toolformer \citep{schick2023toolformer}, Gorilla \citep{patil2024gorilla}, Voyager \citep{wang2023voyager}, RLEF \citep{gehring2025rlef}, and, for website generation specifically, WebGen-Agent's multi-level visual/functional feedback \citep{webgen2025}), together with the thinking-vs-doing analysis of \citet{shen2025thinking}, establish that acting on an environment helps. We contribute the information account of \emph{why}, a coverage principle that predicts \emph{when}, and, crucially, the requirement that the \emph{evaluation} be grounded too. We also compare harness architectures (single-agent loops \citep{trankiela2026singleagent}, sample-and-select \citep{ruan2025iad}, proposer--reviewer) under a matched budget.

\paragraph{Model-as-judge reliability.} LLM- and VLM-as-judge evaluation is now standard \citep{zheng2023judging}, and its failure modes (position, verbosity, and self-preference biases, and broader reward-model fragility \citep{casper2023open}) are documented. Our finding is sharper and, to our knowledge, not previously isolated: for artifacts whose quality is a measurable \emph{physical} property (the geometry of a rendered layout), a screenshot judge is not merely biased but \emph{structurally blind} (the defect is dropped by its observation channel before judgment begins), so it both fails to score the defect and, used as a reviewer, fails to fix it. This is why a deterministic instrument is necessary, not merely preferable, to detect interaction scaling on visual modalities.

\paragraph{Practitioner evidence at scale.} Industrial deployment corroborates the gap our framework targets. \citet{hong2026aicoding} reports a controlled study at ByteDance ($3$ frontier coding models $\times$ $3$ agent frameworks $\times$ $100$ runs) on a single real product feature: functional \emph{correctness} exceeded $80\%$ for every pair, yet ``deliverability'' axes (usability, reliability, maintainability, performance) collapsed to $40$--$60\%$ and rose to ${\sim}80\%$ only after harness ``infrastructure'' was added, while correctness barely moved. This is the signature we formalize: internal scaling saturates the easy-to-see axis while the interaction loop carries the rest. Tellingly, the report had to define a multi-axis quality metric before the gap was visible at all (an industrial analogue of our grounded-evaluation requirement), and its grounded fixes (e.g.\ browser-driven validation before commit) are instrument observations in our taxonomy, not ungrounded self-review.

\paragraph{Information theory.} The data-processing inequality \citep{cover2006elements} underlies the formal version of the internal ceiling (\Cref{app:ceiling}): any post-processing of the model's outputs (reasoning, ungrounded review) cannot increase information about the target, whereas an instrument observation introduces a new channel.

\section{Limitations}
\label{sec:limitations}

Two modalities saturate: video editing is already strong single-shot (the lift is real only on a hardened multi-step suite), and deep research saturates because frontier models know well-documented facts and the judge cannot grade facts it does not have. Deterministic geometry applies only to artifacts with a measurable intended layout; for genuinely flowing content we still rely on a binary rubric, with its VLM-reliability caveat. Our static-form tier is confirmed only on its null prediction (no lift on bugs invisible in form), not yet on a positive one. And because the geometric instrument both supplies the feedback and scores the result, the reported reductions are partly an optimization of the metric itself; we argue in \Cref{sec:grounded_eval} (conservative thresholds, non-trivial magnitude, and the ungrounded-reviewer control that moves the same metric the wrong way) that they reflect real, human-visible defects, but a quantitative human-preference validation remains future work.

\section{Conclusion}
\label{sec:conclusion}

Grounding is the load-bearing variable, and it must hold on both sides of the loop: the feedback that drives revision and the metric that scores it. Reasoning and sampling are internal and hit an information ceiling; interaction escapes it because a grounded instrument imports a real observation each cycle, bounded only by what that instrument can observe. The practical lesson is not ``add more loops'' \citep{swyx2026loopcraft,langchain2026loopeng} but ``ground the loop you add'': wrap a frozen model in a proposer--reviewer harness whose instrument covers the defects that matter, and for visual artifacts measure the rendered DOM, never a screenshot, on both the reviewer and the scorer. Much recent progress on self-improving visual generation is measured with VLM judges that cannot see the defects at issue; grounding the metric is a cheap, deployable correction. Two directions follow: deterministic instruments for modalities beyond layout and execution (audio, 3D, tabular, UI-interaction artifacts), where model judges are likewise blind, and tightening the internal ceiling from an information bound into a quantitative relationship between coverage and achievable quality. Interaction scaling is real, distinct from reasoning and sampling, and predictable from coverage; once you ground the metric, it is plainly visible.

\section*{Acknowledgements}

This paper was produced using Pine Copilot's voice-directed \emph{whisper coding} workflow~\citep{pineai2026whispercoding}, in which the authors specify, discuss, and review the work by voice while a coding agent (Claude Code with Claude Opus 4.8) carries out the planning, coding, experiments, and paper writing.
We thank BSQL Networking for hosting the NVIDIA RTX PRO 6000 GPU.

\bibliographystyle{plainnat}
\bibliography{refs}

\appendix
\section{Formal Statement of the Internal Ceiling}
\label{app:ceiling}

This appendix formalizes the one-sentence argument of \Cref{sec:ceiling}. Let $A^\star$ be the (task-determined) correct artifact and $A$ a candidate. The proposer is a channel from its inputs $\Theta$ (weights) and prompt $X$ to $A$. A \emph{reviewer} that only re-reads $A$ and reasons (ungrounded critique) forms the Markov chain $A^\star \to (\Theta,X) \to A \to R$, where $R$ is its critique. By the data-processing inequality \citep{cover2006elements},
\[
I(A^\star; R)\ \le\ I(A^\star; (\Theta,X)),
\]
so no amount of reasoning or ungrounded review can recover more information about $A^\star$ than is already in $(\Theta,X)$. This bounds reasoning-only scaling. Sampling is bounded differently, but no less tightly: for any finite $N$, best-of-$N$ can only \emph{select} among the $N$ candidates the proposer actually draws, so even an oracle (grounded) verifier cannot return a correct artifact the proposer is too unlikely to sample within the budget. Its practical ceiling is therefore the high-probability part of the proposer's own output distribution, not a new information channel.

Grounded feedback escapes both ceilings by feeding an instrument observation $E=g(A,\,\mathrm{world})$ (a test outcome, a measured bounding box, a search verdict) back into \emph{generation}. Conditioning the next proposal on $E$ lets the proposer synthesize a candidate it would not otherwise have sampled: formally, $I(A^\star;(R,E))$ can exceed $I(A^\star;(\Theta,X))$, because $E$ carries information obtained by running the artifact against reality (the tests encode the specification; the layout engine encodes real rendering semantics the model only approximates). The \emph{coverage} qualification of \Cref{sec:taxonomy} enters here. $E$ helps only to the extent that $g$'s observation is informative about the defects actually present; a linter's $g$ observes form, so on runtime-logic bugs $I(A^\star;E)$ adds nothing actionable, which is what \Cref{fig:typecontrol} measures.

The symmetric claim covers the scorer. A metric $M$ that is itself a model's function of a lossy view of $A$ (a VLM judging a cropped screenshot) cannot certify an improvement that lives in the part of $A$ its channel drops. Grounding the evaluation is therefore a precondition for \emph{detecting} interaction scaling on any modality whose quality is not fully visible to the default judge.

Two cautions. The inequality bounds the \emph{information channel}, not achievable quality directly; we use it to motivate the saturation that \Cref{fig:scaling} then measures, and treat the measurement as the evidence. And grounded interaction is of course still bounded, by what the instrument exposes and by the proposer's ability to act on it.

\section{Detailed Results and Configurations}
\label{app:detailed}
This appendix collects the full tables underlying the figures in the main text, from the four-strategy scaling curves through the distillation ablations.

\subsection{Four-strategy scaling at a matched 20K-token budget}
\begin{table}[H]
\centering
\caption{\textbf{The headline scaling result (3-seed mean $\pm$ SD), underlying \Cref{fig:scaling}.} Pass rate by strategy~$\times$~budget on 15 hard code tasks, Sonnet 4. S, L, and H are averaged over three independent seeds; R was run once (its seed-to-seed variation is small at temp~0), at the sweep's budget partition (a more generous partition lifts R to $86.7\%$; see \Cref{tab:reasoning-baseline}). S's small dip from $B{=}1$K (80.0\%) to $B{=}5$K (77.8\%) is within the $\pm 3.1$\,pp seed noise: pass@$N$ is non-decreasing in $N$ in expectation, but not on every finite sample. Reasoning-only and best-of-$N$ both saturate well below 100\%, while all three feedback-iterating strategies reach $\geq97.8\%$ at $B{=}20$K (L hits 100\% on 2 of 3 seeds, H on all 3, so H is the only variant at a strict 100\% across seeds). H and L tie on pass rate within seed noise; what separates them is token efficiency and reliability (\Cref{fig:efficiency}).}
\label{tab:scaling-curves}
\small
\begin{tabular}{lcccc}
\toprule
Budget $B$ & R (reasoning-only) & S (best-of-$N$) & L (single-agent loop) & H (proposer--reviewer) \\
\midrule
1K   & 60.0\% & 80.0\% $\pm 0.0$\,pp & 57.8\% $\pm 3.1$\,pp & 62.2\% $\pm 3.1$\,pp \\
5K   & 73.3\% & 77.8\% $\pm 3.1$\,pp & \textbf{93.3\% $\pm 0.0$\,pp} & 91.1\% $\pm 3.1$\,pp \\
20K  & 73.3\% & 86.7\% $\pm 0.0$\,pp & 97.8\% $\pm 3.1$\,pp & \textbf{100.0\% $\pm 0.0$\,pp} \\
\midrule
Ceiling & 73.3\% & 86.7\% & $\sim$100\% & \textbf{100\% (zero variance)} \\
\bottomrule
\end{tabular}
\end{table}

\subsection{Reasoning-only at a matched budget}
\begin{table}[H]
\centering
\caption{Reasoning-only at a matched budget vs.\ single-shot and the harness on the 15 hard code tasks. These are single-seed numbers (the reference seed, also used in \Cref{tab:type-controls}); the 3-run aggregate is in \Cref{tab:heldout} ($66.7 \pm 6.7$\% single-shot, $100.0 \pm 0.0$\% reviewed). \emph{Reading the numbers:} single-shot on this seed is $11/15 = 73.3\%$ (the canonical 3-seed mean is $66.7\%$), and the reasoning value shown is the \emph{thinking-heavy} partition ($86.7\%$). Two numeric coincidences of this small suite can mislead: $86.7\%$ here matches the oracle best-of-$N$ ceiling, and the \emph{default}-partition reasoning ceiling ($73.3\%$, \Cref{tab:scaling-curves}) matches this seed's single-shot; neither is the same run. Reasoning closes two-thirds of the harness's gain over single-shot, but stays 6.7\,pp short of the harness even at $1.57\times$ its token budget.}
\label{tab:reasoning-baseline}
\small
\begin{tabular}{lccc}
\toprule
Strategy & Pass rate & Mean tokens & $\Delta$ vs.\ single-shot \\
\midrule
Single-shot                       & 73.3\% (11/15) & 1{,}406  & n/a \\
Reasoning-only (thinking-heavy partition)   & 86.7\% (13/15) & 4{,}137  & $+13.3$\,pp \\
\textbf{Proposer--reviewer harness} & \textbf{93.3\% (14/15)} & \textbf{2{,}643}  & $\mathbf{+20.0}$\,pp \\
\bottomrule
\end{tabular}
\end{table}

\subsection{Feedback-channel control on a fixed code suite}
\begin{table}[H]
\centering
\caption{Feedback-channel controls on 15 hard code tasks (Sonnet 4 proposer and reviewer), underlying \Cref{fig:typecontrol}. Each row is a single seed. ``SS'' is that seed's single-shot pass rate (it varies by row), and ``Reviewed'' is the ceiling after that seed's harness run. The three SS values (10/15 and 11/15) are draws from the same single-shot distribution behind the 3-run mean of $66.7 \pm 6.7$\% (\Cref{tab:heldout}; within $\pm 1$\,SD of 10/15). Since SS varies by row, the comparison that matters is the \emph{reviewed} ceiling: execution reaches 14/15, critique and critique+linter reach 13/15 ($+6.7$\,pp), \emph{and} execution is ${\sim}2.5\times$ more token-efficient ($2{,}643$ vs.\ ${\sim}7{,}100$ tokens/task, 1.53 vs.\ 2.07 iterations). The one task no control row recovers is \texttt{code\_011} (CJK text wrapping); the full harness at $B{=}20$K recovers it on all three seeds (\Cref{tab:scaling-curves}), which is why the headline reviewed rate is a strict $100\%$ while this leaner single-seed control caps at $14/15$.}
\label{tab:type-controls}
\small
\begin{tabular}{lcccc}
\toprule
Feedback channel & Reviewer sees & SS & Reviewed & Iters \\
\midrule
None (single-shot)             & n/a                                  & 11/15 & n/a            & n/a \\
Ungrounded critique            & code only                            & 10/15 & 13/15          & 2.07 \\
$+$ linter (static form)       & code $+$ \texttt{ruff} output        & 11/15 & 13/15          & 2.07 \\
\textbf{$+$ execution}         & code $+$ test stderr/traceback       & 11/15 & \textbf{14/15} & \textbf{1.53} \\
\bottomrule
\end{tabular}
\end{table}

\subsection{Single-agent loop vs.\ proposer--reviewer harness}
\begin{table}[H]
\centering
\caption{Three interaction-strategy variants at $B{=}20$K on the 15 hard code tasks, all using execution feedback; underlying \Cref{fig:efficiency}. Pass rate is within seed noise (sign-test $p{>}0.6$ on the H--L comparison). The harness wins on \emph{token efficiency} (1{,}029 vs.\ 1{,}431 L vs.\ 1{,}416 IAD mean output tokens) and on \emph{seed variance} (zero across three seeds for H, while L hit 100\% in 2 of 3 seeds and IAD was run for one seed only).}
\label{tab:LvsH}
\small
\begin{tabular}{lccc}
\toprule
Strategy & Pass rate & Mean tokens & Notes \\
\midrule
L (single-agent loop)              & 97.8\% $\pm 3.1$\,pp (3 seeds)           & 1{,}431          & 1 task in 1 seed \\
IAD (oracle, K=3, T=0.7)           & 100.0\% (1 seed)                         & 1{,}416          & $+38$\% tokens vs.\ H \\
\textbf{H (proposer--reviewer)}    & \textbf{100.0\% $\pm 0.0$\,pp (3 seeds)} & \textbf{1{,}029} & zero seed variance \\
\bottomrule
\end{tabular}
\end{table}

\subsection{Deterministic geometric-feedback statistics}
\begin{table}[H]
\centering\small
\caption{Full statistics for \Cref{fig:geomreduction}: deterministic geometric-feedback results, one identical configuration across four visual modalities (Sonnet~4, $T{=}0$, $3$ seeds, alignment-inclusive reward). SS/Rev.\ are mean defects per artifact, single-shot vs.\ reviewed; $\Delta$ is the mean defect reduction; CI is a paired bootstrap 95\% interval on $\Delta\%$ ($10$k resamples); $p$ is a two-sided paired sign test over decisive (task, seed) pairs. Web is summed over $1920/375$ widths, animations over sampled frames.}
\label{tab:geometric}
\begin{tabular}{lcccccc}
\toprule
Modality & $n$ & SS & Rev. & $\Delta$ & 95\% CI & impr./regr.\ ($p$) \\
\midrule
Academic figures (20) & 60 & $0.57$ & $0.15$ & $-74\%$ & $[52,89]\%$ & $17/2$\;($7{\times}10^{-4}$) \\
Dense slides (12)     & 36 & $1.25$ & $0.33$ & $-73\%$ & $[45,93]\%$ & $13/1$\;($1.8{\times}10^{-3}$) \\
Web pages (20)        & 60 & $16.1$ & $8.5$  & $-47\%$ & $[30,62]\%$ & $33/2$\;($4{\times}10^{-8}$) \\
Animations (20)       & 60 & $16.9$ & $10.2$ & $-40\%$ & $[10,64]\%$ & $34/7$\;($3{\times}10^{-5}$) \\
\bottomrule
\end{tabular}
\end{table}

\subsection{Cross-model replication (code)}
\begin{table}[H]
\centering
\caption{Cross-model replication on the same 15 hard code tasks, each family over \emph{three} independent on-policy seeds at $T{=}0.7$ (Claude is the multi-run reference from \Cref{tab:heldout}); underlying \Cref{fig:crossmodel-code}. The reviewed ceiling has zero seed variance for all three families, and the harness lift survives across Anthropic, Alibaba, and OpenAI. Single-shot per-seed pass rates: Qwen $66.7/73.3/73.3\%$, GPT-5 $86.7/80.0/73.3\%$.}
\label{tab:crossmodel}
\small
\begin{tabular}{lccc}
\toprule
Model & Single-shot pass & Reviewed pass & $\Delta$ (lift) \\
\midrule
Claude Sonnet 4 (3 seeds)            & $66.7 \pm 6.7$\%   & $\mathbf{100.0 \pm 0.0}$\%      & $\mathbf{+33.3}$\,pp \\
Qwen3-235B-Instruct-2507 (3 seeds)   & $71.1 \pm 3.8$\%   & $\mathbf{93.3 \pm 0.0}$\%       & $\mathbf{+22.2}$\,pp \\
GPT-5 (3 seeds)                      & $80.0 \pm 6.7$\%   & $\mathbf{100.0 \pm 0.0}$\%      & $\mathbf{+20.0}$\,pp \\
\bottomrule
\end{tabular}
\end{table}

\subsection{Held-out generalization (code)}
\begin{table}[H]
\centering
\caption{Held-out generalization on 32 newly constructed code tasks (zero overlap with the 15-task development set). The harness recovers 3/3 single-shot failures and regresses zero passing tasks. The smaller absolute $\Delta$ is a baseline-compression effect (single-shot is 90.6\% on the held-out set, leaving at most 9.4\,pp headroom).}
\label{tab:heldout}
\small
\begin{tabular}{lccccc}
\toprule
Split & $N$ & Single-shot & Reviewed & $\Delta$ & SS-fail fixes\,/\,regr. \\
\midrule
Development (3-run mean) & 15 & $66.7 \pm 6.7$\% & $100.0 \pm 0.0$\% & $+33.3$\,pp & 5/5 / 0 \\
\textbf{Held-out v2}     & 32 & $90.6$\% & $\mathbf{100.0}$\% & $\mathbf{+9.4}$\,pp & $\mathbf{3/3}$ / $\mathbf{0}$ \\
\bottomrule
\end{tabular}
\end{table}

\subsection{Budget-allocation simplex}
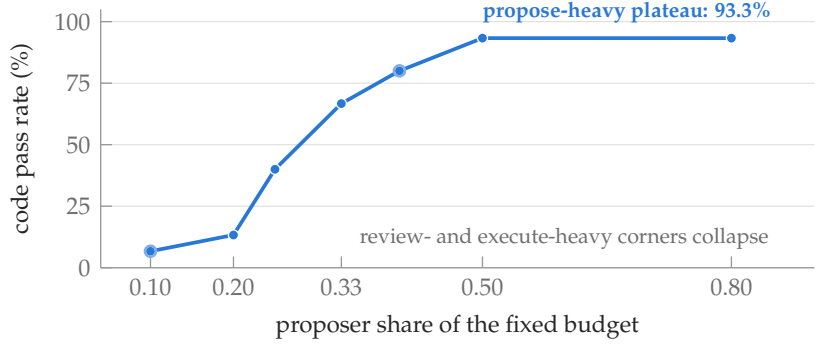
\begin{figure}[H]
\centering
\begin{tikzpicture}
\begin{axis}[width=0.7\linewidth,height=5.0cm,
  xlabel={proposer share of the fixed budget},ylabel={code pass rate (\%)},
  xmin=0.04,xmax=0.9,ymin=0,ymax=105,clip=false,
  xtick={0.1,0.2,0.33,0.5,0.8},
  xticklabels={0.10,0.20,0.33,0.50,0.80},
  ytick={0,25,50,75,100},ymajorgrids,
  axis x line*=bottom,axis y line*=left]
  \addplot[cGnd,line width=1.4pt,mark=*,mark size=1.9,
    mark options={fill=cGnd,draw=white,line width=0.5pt}]
    coordinates {(0.10,6.7)(0.20,13.3)(0.25,40.0)(0.33,66.7)(0.40,80.0)(0.50,93.3)(0.80,93.3)};
  \addplot[only marks,mark=o,mark size=2.1,cGnd!60,line width=0.9pt]
    coordinates {(0.10,6.7)(0.40,80.0)};
  \node[font=\scriptsize,cMut,anchor=west,align=left] at (axis cs:0.34,12)
    {review- and execute-heavy corners collapse};
  \node[font=\scriptsize\bfseries,cGnd,anchor=south east,align=right] at (axis cs:0.86,95.5)
    {propose-heavy plateau: 93.3\%};
\end{axis}
\end{tikzpicture}
\caption{\textbf{Pass rate is monotone in the proposer's budget share.} Nine allocations of a fixed $10$K-token budget across proposer / execution / reviewer, on the hard code suite (open markers: a second allocation with the same proposer share but a different execution/review split, showing that the proposer share alone predicts the outcome). The spread across the sweep is $86.6$\,pp; full simplex in \Cref{tab:allocation}.}
\label{fig:allocation}
\end{figure}

Sweeping how a fixed token budget is split between proposer, execution, and reviewer produces an enormous spread in pass rate (\Cref{fig:allocation}), rising monotonically with the proposer's share: review-heavy corners collapse almost to zero, while propose-heavy splits plateau at the top. The best split is \emph{propose-heavy} (give most of the budget to generation, reserving just enough for the reviewer to locate defects); extra compute is better spent on more samples than on deeper iteration past the early-stop point.

\begin{table}[H]
\centering
\caption{Budget allocation simplex on 15 hard code tasks, $B = 10$K total; underlying \Cref{fig:allocation}. Pass rate is monotone in the proposer's per-call cap, and review-heavy corners (low $b_1$) collapse. The 9-point sweep yields an 86.6\,pp spread.}
\label{tab:allocation}
\small
\begin{tabular}{lccccc}
\toprule
Allocation & $b_1$ (prop) & $b_2$ (exec) & $b_3$ (rev) & Pass rate & Mean tokens \\
\midrule
\textbf{A (propose-heavy)} & 0.80 & 0.10 & 0.10 & \textbf{93.3\% (14/15)} & 3{,}101 \\
\textbf{G (prop-dominant)} & 0.50 & 0.25 & 0.25 & \textbf{93.3\% (14/15)} & 2{,}697 \\
D (prop+exec)              & 0.40 & 0.40 & 0.20 & 80.0\% (12/15) & 4{,}025 \\
E (prop+review)            & 0.40 & 0.20 & 0.40 & 80.0\% (12/15) & 4{,}388 \\
I (equal)                  & 0.33 & 0.34 & 0.33 & 66.7\% (10/15) & 4{,}845 \\
H (review-dominant)        & 0.25 & 0.25 & 0.50 & 40.0\% (6/15)  & 7{,}485 \\
F (exec+review)            & 0.20 & 0.40 & 0.40 & 13.3\% (2/15)  & 8{,}771 \\
B (execute-heavy)          & 0.10 & 0.80 & 0.10 & \phantom{0}6.7\% (1/15)  & 9{,}383 \\
C (review-heavy)           & 0.10 & 0.10 & 0.80 & \phantom{0}6.7\% (1/15)  & 9{,}383 \\
\midrule
\multicolumn{4}{r}{Spread:} & \textbf{86.6\,pp} & \\
\bottomrule
\end{tabular}
\end{table}

\subsection{Token ROI by modality}
\begin{table}[H]
\centering
\caption{Per-task token cost and quality return-on-investment for the harness, averaged across three on-policy runs. ``Extra tokens'' is reviewed-minus-single-shot per task. Tokens per 0.01 quality is extra tokens divided by (mean reviewed quality $-$ mean single-shot quality)$\cdot 100$. Code dominates by $100\times$.}
\label{tab:harness-roi}
\small
\begin{tabular}{lrrrr}
\toprule
Modality & Single-shot tokens & Reviewed tokens & Extra tokens & Tokens / 0.01 quality \\
\midrule
Code        & 3{,}336  & 4{,}575  & 1{,}239  & \textbf{37} \\
Video       & 3{,}808  & 20{,}584 & 16{,}776 & 359 \\
Research    & 14{,}116 & 27{,}793 & 13{,}677 & 2{,}415 \\
Webpages    & 14{,}060 & 80{,}148 & 66{,}088 & 5{,}128 \\
Animations  & 25{,}491 & 89{,}009 & 63{,}518 & 4{,}331 \\
Slides      & 10{,}274 & 36{,}956 & 26{,}682 & 4{,}447 \\
\bottomrule
\end{tabular}
\end{table}

\subsection{Distillation: headline}
\begin{table}[H]
\centering
\caption{Distilled 8B student on the 18-task hard held-out suite (teacher's pre-curated single-shot failures, so teacher single-shot is 0/18 by construction). The student lifts judge-keep from 0\% to 44\% at pass@1 and 56\% at pass@3. The $0.51\times$ mean@1 capability ratio against the teacher is established on a separate 44-task out-of-distribution suite (\Cref{tab:distill-ood}), where teacher single-shot is non-degenerate.}
\label{tab:distill-headline}
\small
\begin{tabular}{lcc}
\toprule
Setup & Compute & Judge-keep \\
\midrule
1 sample $\times$ 5 turns                                  & $1\times$ & 44\% (8/18) \\
1 sample $\times$ 8 turns                                  & $1.6\times$ & 28\% (regressed) \\
\textbf{3 samples $\times$ 5 turns (pass@3)}               & $3\times$ & \textbf{56\% (10/18)} \\
\bottomrule
\end{tabular}
\end{table}

\subsection{Distillation: out-of-distribution capability ratio}
\begin{table}[H]
\centering
\caption{Distilled 8B student (V3 $+$ force-close) on a 44-task out-of-distribution suite where the teacher's single-shot is non-degenerate (teacher $20/44$). Capability ratio is the student's judge-keep divided by the teacher's $20/44$. Two sampling seeds are used, and pass@2 is the union over both. This is the suite that establishes the ${\sim}0.51\times$ mean@1 / $0.70\times$ pass@2 ratios quoted in \Cref{sec:distill} and \Cref{fig:distill}.}
\label{tab:distill-ood}
\small
\begin{tabular}{lcccc}
\toprule
Metric & sample 1 & sample 2 & mean@1 & pass@2 \\
\midrule
Judge-keep                              & $9/44$ ($20\%$) & $11/44$ ($25\%$) & $23\%$        & $\mathbf{32\%}$ \\
Capability ratio (vs teacher $20/44$)   & $0.45\times$    & $0.55\times$     & $0.51\times$  & $\mathbf{0.70\times}$ \\
\bottomrule
\end{tabular}
\end{table}

\subsection{Distillation: capability U-curve over force-close budget}
\begin{table}[H]
\centering
\caption{Reasoning-cap U-curve on the 18-task hard held-out suite. SFT recipe sweep on Qwen3-VL-8B-Instruct (37 judge-kept teacher traces, same inference protocol). The optimum is at the student's coherence horizon, not the teacher's.}
\label{tab:distill-ucurve}
\begin{tabular}{lccc}
\toprule
Variant & Reasoning cap (chars) & Judge-keep & Identical-retry \\
\midrule
V1 (no reasoning)     & 0    & 6\%  & 67\% \\
V4                    & 1000 & 11\% & 35\% \\
V2                    & 3000 & 17\% & 32\% \\
\textbf{V3 (chosen)}  & \textbf{1500} & \textbf{44\%} & 10\% \\
\bottomrule
\end{tabular}
\end{table}

\subsection{Distillation: data-scale}
\begin{table}[H]
\centering
\caption{Three independent attempts to scale the SFT corpus past V3's 37 examples. All regressed. Quality dominates quantity at this scale.}
\label{tab:distill-datascale}
\begin{tabular}{lccc}
\toprule
Variant & Composition & $n$ examples & Judge-keep \\
\midrule
\textbf{V3 (chosen)} & 235B teacher, judge-kept       & 37 & \textbf{44\%} \\
V5                   & V3 + 13 judge-rejected (235B)  & 50 & 33\% \\
V6                   & V3 + 20 judge-kept (30B teacher) & 57 & 33\% \\
RFT v1               & Self-rolled, judge-filtered    & 61 & 39\% \\
\bottomrule
\end{tabular}
\end{table}

\subsection{Distillation: RFT regresses pass@3}
\begin{table}[H]
\centering
\caption{RFT polishes consistency at the cost of pass@$k$; underlying \Cref{fig:variance}. Per-turn metrics improve, per-task variance collapses, and pass@3 regresses by 17\,pp. For deployment with sampling, the model's variance \emph{is} the inference-scaling budget. \textit{Note on notation.} \textbf{mean@1} is the expected pass rate of a single random sample, computed as the mean accuracy across the three sampling seeds (one $T{=}0$ greedy + two $T{=}0.7$). This is the natural variance-sensitive metric and differs from the greedy single-seed \emph{pass@1}=44\% reported in \Cref{tab:distill-headline}, which is just seed~1's accuracy. \emph{pass@2} and \emph{pass@3} are union pass rates over the first 2 / all 3 seeds.}
\label{tab:distill-rft}
\small
\begin{tabular}{lcccccc}
\toprule
& \multicolumn{3}{c}{Consistency metrics} & \multicolumn{3}{c}{Sampling-aware judge-keep} \\
\cmidrule(lr){2-4} \cmidrule(lr){5-7}
& id-retry $\downarrow$ & addresses-rev $\uparrow$ & no-artifact $\downarrow$ & mean@1 & pass@2 & pass@3 \\
\midrule
\textbf{V3 (chosen)} & 9\%  & 83\% & 2  & \textbf{31\%} & \textbf{50\%} & \textbf{56\%} \\
RFT v1               & \textbf{8\%}  & \textbf{89\%} & \textbf{1}  & 28\%   & 39\%   & 39\% \\
\bottomrule
\end{tabular}
\end{table}

\subsection{Per-task pass@$k$ for the 8B markdown student}
\begin{table}[H]
\centering
\caption{Per-task judge-keep outcome for the 8B markdown student on the 18-task hard held-out suite. $\checkmark$ = judge-kept, $\cdot$ = judge-rejected.}
\label{tab:per-task-md}
\small
\begin{tabular}{llccccc}
\toprule
Category & Task & seed 1 ($T{=}0$) & seed 2 ($T{=}0.7$) & seed 3 ($T{=}0.7$) & pass@3 \\
\midrule
\multirow{5}{*}{webpages} & web\_002 & $\checkmark$ & $\cdot$ & $\cdot$ & $\checkmark$ \\
& web\_003 & $\checkmark$ & $\cdot$ & $\checkmark$ & $\checkmark$ \\
& web\_005 & $\cdot$ & $\cdot$ & $\cdot$ & $\cdot$ \\
& web\_008 & $\cdot$ & $\cdot$ & $\cdot$ & $\cdot$ \\
& web\_012 & $\cdot$ & $\cdot$ & $\cdot$ & $\cdot$ \\
\midrule
\multirow{8}{*}{slides} & slide\_001 & $\cdot$ & $\cdot$ & $\cdot$ & $\cdot$ \\
& slide\_006 & $\cdot$ & $\cdot$ & $\cdot$ & $\cdot$ \\
& slide\_007 & $\cdot$ & $\cdot$ & $\cdot$ & $\cdot$ \\
& slide\_008 & $\checkmark$ & $\cdot$ & $\cdot$ & $\checkmark$ \\
& slide\_010 & $\cdot$ & $\cdot$ & $\cdot$ & $\cdot$ \\
& slide\_014 & $\checkmark$ & $\checkmark$ & $\checkmark$ & $\checkmark$ \\
& slide\_018 & $\checkmark$ & $\checkmark$ & $\cdot$ & $\checkmark$ \\
& slide\_020 & $\cdot$ & $\checkmark$ & $\cdot$ & $\checkmark$ \\
\midrule
\multirow{5}{*}{code} & code\_h001 & $\checkmark$ & $\cdot$ & $\cdot$ & $\checkmark$ \\
& code\_h002 & $\checkmark$ & $\checkmark$ & $\cdot$ & $\checkmark$ \\
& code\_h003 & $\cdot$ & $\cdot$ & $\cdot$ & $\cdot$ \\
& code\_h004 & $\cdot$ & $\cdot$ & $\checkmark$ & $\checkmark$ \\
& code\_h005 & $\checkmark$ & $\checkmark$ & $\checkmark$ & $\checkmark$ \\
\midrule
\multicolumn{2}{r}{Total kept:} & 8/18 & 5/18 & 4/18 & \textbf{10/18 (56\%)} \\
\bottomrule
\end{tabular}
\end{table}

\end{document}